\documentclass[letterpaper]{article} 
\usepackage{aaai2026}  
\usepackage{times}  
\usepackage{helvet}  
\usepackage{courier}  
\usepackage[hyphens]{url}  
\usepackage{graphicx} 
\usepackage{natbib}  
\usepackage{caption} 
\usepackage{algorithm}
\usepackage{algorithmic}
\usepackage{amsfonts}
\usepackage{subcaption}  

\usepackage{times}
\usepackage{soul}
\usepackage{url}
\usepackage{amsmath}
\usepackage{amsthm}
\usepackage{booktabs}
\usepackage{algorithm}
\usepackage{algorithmic}
\usepackage[switch]{lineno}

\usepackage{booktabs}
\usepackage{threeparttable}
\usepackage{makecell}
\usepackage{multirow,color}
 
\newtheorem{theorem}{Theorem}

\newtheorem{definition}{Definition}

\usepackage{newfloat}
\usepackage{listings}
\DeclareCaptionStyle{ruled}{labelfont=normalfont,labelsep=colon,strut=off} 
\floatstyle{ruled}
\newfloat{listing}{tb}{lst}{}
\floatname{listing}{Listing}
\title{Dual-View Inference Attack: Machine Unlearning Amplifies Privacy Exposure}
\author {
Lulu Xue\textsuperscript{\rm1},
Shengshan Hu\textsuperscript{\rm1},
Linqiang Qian\textsuperscript{\rm2},
Peijin Guo\textsuperscript{\rm1},
Yechao Zhang\textsuperscript{\rm3},
Minghui Li\textsuperscript{\rm4} \thanks{Corresponding Author.},
Yanjun Zhang\textsuperscript{\rm5},
Dayong Ye\textsuperscript{\rm6},
 Leo Yu Zhang\textsuperscript{\rm7}
}
\affiliations {
    \textsuperscript{\rm 1} School of Cyber Science and Engineering, Huazhong University of Science and Technology\\
    \textsuperscript{\rm 2}Institute for Network Sciences and Cyberspace, Tsinghua University\\
 \textsuperscript{\rm 3}  College of Computing and Data Science, Nanyang Technological University\\
 \textsuperscript{\rm 4} School of Software Engineering, Huazhong University of Science and Technology\\
    \textsuperscript{\rm 5}School of Computer Science, University of Technology Sydney\\
       \textsuperscript{\rm 6} Faculty of Data Science, City University of Macau\\
    \textsuperscript{\rm 7} School of Information and Communication Technology, Griffith University\\
$\{$lluxue,hushengshan,gpj,minghuili$\}$@hust.edu.cn, qlq25@mails.tsinghua.edu.cn,yech.zhang@gmail.com,Yanjun.Zhang@uts.edu.au,dayongye@outlook.com, leo.zhang@griffith.edu.au}

\usepackage{bibentry}
\begin{document}
\maketitle
\begin{abstract}
Machine unlearning is a newly popularized technique for removing specific training data from a trained model, enabling it to comply with data deletion requests. While it protects the rights of users requesting unlearning, it also introduces new privacy risks. Prior works have primarily focused on the privacy of data that has been unlearned, while the risks to retained data remain largely unexplored. 
To address this gap, we focus on the privacy risks of retained data and, for the first time, reveal the vulnerabilities introduced by machine unlearning under the dual-view setting, where an adversary can query both the original and the unlearned models. From an information-theoretic perspective, we introduce the concept of {privacy knowledge gain} and demonstrate that the dual-view setting allows adversaries to obtain more information than querying either model alone, thereby amplifying privacy leakage.  To effectively demonstrate this threat, we propose DVIA, a Dual-View Inference Attack, which extracts membership information on retained data using black-box queries to both models. DVIA eliminates the need to train an attack model and employs a lightweight likelihood ratio inference module for efficient inference. Experiments across different datasets and model architectures validate the effectiveness of DVIA and highlight the privacy risks inherent in the dual-view setting.
\end{abstract}
\section{Introduction}
\begin{figure}[t!]
    \centering
    \includegraphics[width=8.3cm]{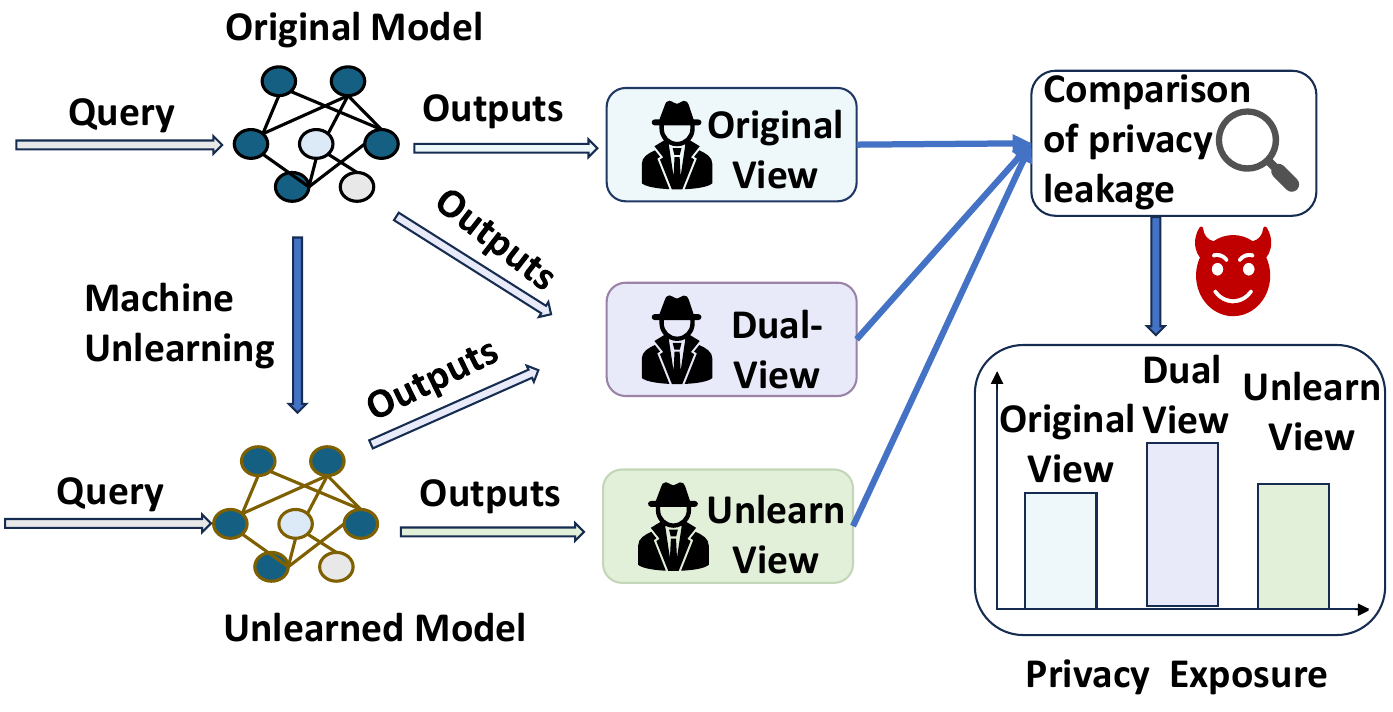}
\caption{An illustration of the dual-view setting, showing that accessing two models can result in greater privacy leakage than accessing a single model alone.
}
    \label{fig:frame}
\end{figure}
Machine unlearning~\cite{MU,GA1} is an advanced technique designed to remove the information of specific training data points from the target model.
While machine unlearning is often considered a privacy-preserving technique because it allows users to remove their private data from a trained model~\cite{privacy1,privacy2,privacy3}, it also introduces new privacy risks. Several studies have examined the privacy leakage of forgotten data during the unlearning process. For example, \cite{recon1,recon2} show that adversaries may be able to partially reconstruct deleted samples, while \cite{U_MIA,u-label} focus on inferring which specific data points have been unlearned. These studies typically treat users of unlearning APIs as privacy victims, but they overlook a critical threat: \textbf{such users may also exploit the unlearning interface as an attack vector to compromise the privacy of retained data (non-forgotten data).}

This threat receives limited attention. Although \cite{carlini2022privacy} introduce the "privacy onion effect," showing that removing certain data may inadvertently increase the privacy risk of retained samples, and \cite{U_LIRA} point out that unlearned models may overfit more severely to the retained data, these analyses focus solely on the unlearned model and remain constrained to the traditional single-model inference setting. They fail to consider a more realistic and concerning scenario in which users who initiate unlearning requests typically access the models before and after unlearning, referred to as the original model and the unlearned model, respectively.
This access is a reasonable requirement, as users need to verify whether their data are successfully removed~\cite{xu2024really}. However, it also introduces a new privacy attack surface against retained data. In this work, we uncover this previously overlooked risk dimension and define this setting as the dual-view scenario, where the two "views" correspond to access to the original and the unlearned models, as illustrated in Fig.~\ref{fig:frame}. Our analysis shows that in this setting, attackers leverage the responses from both models to extract richer membership information and significantly enhance inference on retained data.

Specifically, we adopt an information-theoretic perspective to analyze the additional information an attacker can obtain when accessing both the original and unlearned models, compared to accessing a single model. We define this additional information as the \textit{Privacy Knowledge Gain} (Gain). A positive Gain indicates a reduction in the adversary's uncertainty about the membership status of target data, thereby implying increased privacy leakage. We further analyze the underlying cause of a positive Gain and find that it results from the asymmetric impact of unlearning on model predictions. When a set of data is removed, the model's predictions for non-forgotten (member) data remain largely stable due to memorization, whereas its predictions for unseen (non-member) data are more likely to change.

This asymmetry is especially exploitable in the {dual-view} setting, where adversaries compare the outputs of the original and unlearned models to more effectively infer membership, thereby amplifying the privacy risks of machine unlearning.
Building on this insight, we propose \textbf{DVIA} (\textbf{D}ual-\textbf{V}iew \textbf{I}nference \textbf{A}ttack), the first membership inference attack for retained dataset tailored to the dual-view setting. DVIA leverages the asymmetric impact introduced by unlearning to improve inference performance.
It introduces a new metric, {Unlearning Confidence Difference} (UCD), which quantifies the impact of unlearning on target dataset. In addition, DVIA incorporates a likelihood ratio–based inference module that requires no training and uses a fixed threshold, enabling efficient and accurate inference without auxiliary models or threshold tuning.

In summary, our contributions are as follows: 
1) We reveal a previously overlooked privacy threat in machine unlearning, where users initiating unlearning requests gain a significant advantage in inferring the membership of retained data by accessing both the original and the unlearned models.
2) We formalize this threat as a dual-view setting, analyze its amplified privacy risks, and propose DVIA, the first membership inference attack specifically designed for retained data in this setting.
3) We conduct extensive experiments across diverse datasets and model architectures, showing that DVIA achieves strong performance in uncovering membership information under unlearning scenarios.

\section{Related Works}\label{sec:relatedwork}
\subsection{Machine Unlearning} 
Machine unlearning~\cite{M_survey3,eu2,eu3} is the process of initiating a reverse learning task to eliminate the information of specific data points on the target model. 
Machine unlearning can be divided into exact unlearning~\cite{e1,eu1} and approximate unlearning~\cite{MU8,GA1}. Exact unlearning involves training model parameters from scratch using the retain dataset, and the retrain method, which retains the original settings for retraining, offers the best unlearning results. While exact unlearning ensures effective data removal~\cite{sparse,Salun}, it comes at a high computational cost~\cite{l1,l2,l3,l4}. To address this challenge, many approximate unlearning algorithms have been proposed~\cite{IU2,GA1,GA2,Salun,sparse,IU3}. Although these methods introduce some inaccuracies during the unlearning process, they demonstrate the feasibility of quickly eliminating the influence of forgotten data on model parameters.
\subsection{Membership Inference Attacks}
Membership inference attacks (MIAs)~\cite{MIA1,lira,amplifying} aim to determine whether a given sample was part of a model’s training set. Most existing approaches rely on shadow models: adversaries train multiple models on auxiliary data to simulate the target model’s behavior, then use the resulting labeled data to train a binary classifier~\cite{MIA1,amplifying} or to select a decision threshold~\cite{gap,label_only}. These methods often require careful design and threshold tuning.
Recently, \cite{lira} proposes a likelihood ratio–based approach that avoids explicit attack model training and complex threshold selection. However, it conducts inference on a per-sample basis and requires generating multiple shadow models and indicators for each target sample, leading to high computational overhead.

\noindent \textbf{Membership inference in machine unlearning.} 
Membership inference is commonly used as a tool to evaluate the effectiveness of forgetting~\cite{sparse,Salun}.  
Recent studies~\cite{U_MIA,U_LIRA} also explore its use in exposing privacy risks introduced by unlearning, focusing on inferring deleted data.  
These studies typically treat users of unlearning APIs as victims, aiming to recover forgotten samples.  However, prior studies~\cite{carlini2022privacy,U_LIRA} suggest that unlearning may amplify privacy leakage, yet they do not propose dedicated attacks for this setting and overlook the increased information advantage that unlearning grants to adversaries.
\section{Privacy Analysis}
\subsection{Threat Model}
We consider a user invoking the unlearning API as a curious adversary aiming to infer membership. The adversary requests the removal of a forgotten dataset \(D_f\) and queries both the original model \(\mathbf{\theta}^o\) and the unlearned model \(\mathbf{\theta}^u\), forming a {dual-view setting}. This setting is realistic, since users can often access both models to confirm data removal~\cite{xu2024really,xue2025towards}.
Using this access, the adversary aims to infer the membership status of samples in a target dataset \(D_t\), where samples from the retained set \(D_{\text{retain}}\) are considered members and those from the validation set \(D_{\text{val}}\) are non-members. We refer to this as the \emph{non-forgetting inference} scenario.
Following~\cite{U_MIA,U_LIRA}, we assume a black-box setting where the adversary knows 
the data distribution $\mathbb{D}$, as well as the training and unlearning procedures 
and the model architecture, but not the model parameters. 
\subsection{Privacy Knowledge Gain}
\label{sec:gain}
 To analyze the privacy risks introduced by this dual-view, we adopt an information-theoretic perspective and define the adversary’s improved privacy knowledge as \emph{privacy knowledge gain}. 
 The specific definition is as follows.
\begin{definition}[Privacy Knowledge Gain] 
Let \( s_o(D_t) \) and \( s_u(D_t) \) denote the observable behaviors of the original and unlearned models on the target dataset \( D_t \), and let \( s(D_t) \in \{s_o(D_t), s_u(D_t)\} \) denote a single-view observation. The \emph{privacy knowledge gain} measures the additional membership information revealed by dual-view access, defined as:
\[
\textnormal{Gain}(D_t)= I(M_{D_t}; s_o(D_t), s_u(D_t)) - I(M_{D_t}; s(D_t)),
\]
where \( M_{D_t} \in \{0,1\}^{|D_t|} \) denotes the membership status of each sample in \( D_t \), with 1 indicating member data. \( I(\cdot ; \cdot) \) denotes the mutual information~\cite{shannon1948mathematical}, which measures the reduction in uncertainty of one variable given knowledge of another.
\end{definition}
When \( \text{Gain}(D_t) > 0 \), it indicates that the adversary has acquired additional information about the membership status of samples in \( D_t \), reducing uncertainty and enabling more effective privacy inference. Our goal is to analyze the conditions under which \( \text{Gain}>0 \) holds in the {dual-view} setting, in order to reveal how this setting amplifies privacy leakage risks. Specifically, we investigate the necessary and sufficient conditions for \(\text{Gain}(D_t) > 0 \) under machine unlearning, and validate them with practical examples. To facilitate this analysis, we first define the behavioral impact of machine unlearning on the \( D_t \) as follows:
\begin{definition}
The behavioral impact of machine unlearning on the target dataset \( D_t \) is defined as
\[
\delta(D_t) = 1 - \text{sim}\big(s_o(D_t), s_u(D_t)\big),
\]
where \( s_o(D_t) \) and \( s_u(D_t) \) denote the observable behaviors of the original and unlearned models on \( D_t \), respectively, and \( \text{sim}(\cdot, \cdot) \) is a similarity function with larger values indicating higher behavioral closeness. \end{definition}
{Intuitively, if \(\delta(D_t)\) is independent of the membership status of \(D_t\), this implies that the behavioral change between the two models does not statistically differ for members and non-members. Therefore, the unlearned model does not expose any additional membership information beyond what is already present in the original model. Conversely, if \(\delta(D_t)\) exhibits statistical differences conditioned on membership status, then it must encode additional information about whether a sample belongs to the retained dataset. This additional signal can be exploited by an adversary to improve membership inference, thereby resulting in a positive privacy gain.}
Based on this definition, we formalize the notion of privacy knowledge gain under membership conditions in the following theorem.
\begin{theorem}
For the target dataset \( D_t \), The privacy knowledge gain satisfies:
\begin{align*}
\textnormal{Gain}(D_t) > 0 
\quad &\text{if and only if} \\
\mathbb{P}(\delta(D_t) \mid M_{D_t} = 1) 
&\ne 
\mathbb{P}(\delta(D_t) \mid M_{D_t} = 0),
\end{align*}
where \( \delta(D_t)\) is the behavioral impact for \( D_t \),
\( M_{D_t} \) is the membership status of \( D_t \), and \( \mathbb{P}(\cdot \mid \cdot) \) denotes conditional probability.
\label{theo:t1}
\end{theorem}
The proof is provided in the appendix. 
According to Theorem~\ref{theo:t1}, the privacy knowledge of the target dataset \( D_t \) is amplified by machine unlearning if the observable behavioral impact of unlearning on member and non-member data differs. To quantify this behavioral impact, we adopt the influence function proposed in~\cite{if}, which estimates how adding a training point \( \mathbf{z} \) affects the model's prediction on a given test point \( \mathbf{z}_{\text{test}} \). The influence is computed as:
\begin{equation}
\mathcal{I}_{\text{IF}}( \mathbf{z} , \mathbf{z}_{\text{test}} ) = -\nabla_{\boldsymbol{\theta}} \mathcal{L}(\mathbf{z}_{\text{test}}, \boldsymbol{\theta})^\top H_{\boldsymbol{\theta}}^{-1} \nabla_{\boldsymbol{\theta}} \mathcal{L}(\mathbf{z}, \boldsymbol{\theta}),
\label{Eq:inf}
\end{equation}
where \( \boldsymbol{\theta} \) denotes the model parameters, \( \mathcal{L} \) is the loss function, and \( H_{\boldsymbol{\theta}} \) is the Hessian matrix of second-order derivatives of \(\mathcal{L}\) with respect to \( \boldsymbol{\theta} \).
In the scenario of removing data points from the training set, we use \( -\mathcal{I} _{IF} \) to measure the impact of unlearning. Specifically, \(-\mathcal{I}_{\text{IF}}(D_f, \mathbf{z}_{\text{test}})\) measures the behavioral change of the model on \(\mathbf{z}_{\text{test}}\) resulting from unlearning the dataset \(D_f\), where \(\mathbf{z}_{\text{test}}\) denotes an arbitrary sample from the target dataset \(D_t\).
 Due to the high computational cost of directly calculating the Hessian matrix, we approximate it using the woodfisher information matrix~\cite{woodfisher} in this paper.
 
\begin{figure}[t!]
    \centering
    \includegraphics[width=8cm]{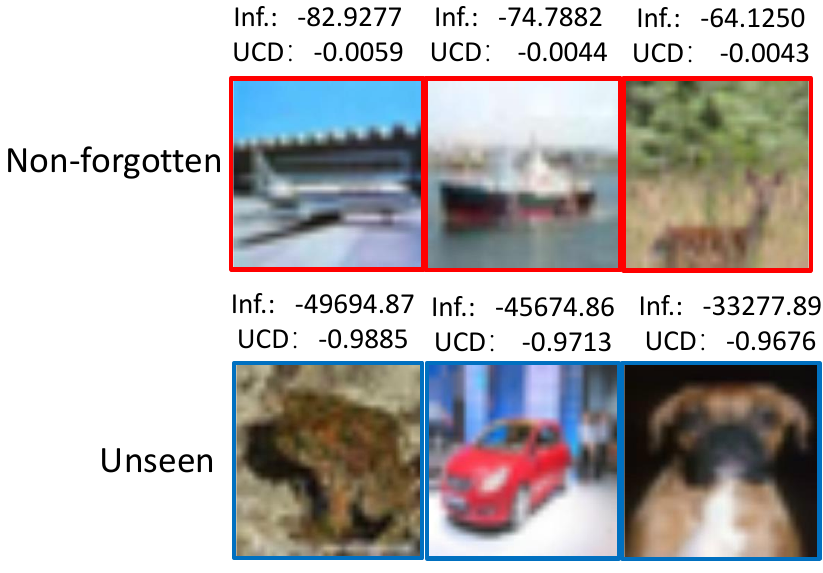}
\caption{A depiction of the unlearning influence on the target model’s non-forgotten and unseen data. ``Inf." refers to the influence score, and UCD is the metric we propose to measure the behavioral impact  in black-box settings.}
    \label{fig:inf}

\end{figure}

Based on the influence function, we perform a straightforward experiment by training a convolutional neural network model, ALLCNN~\cite{ALLCNN}, on the CIFAR10 dataset and subsequently unlearning 5\% of the data points. We sample 100 points from the non-forgotten set and validation set respectively to calculate their influence scores, and display the top 3 images with the largest absolute influence scores, as shown in Figure~\ref{fig:inf}.  
{It can be observed that machine unlearning typically has a smaller impact on non-forgotten data points, while exerting a larger influence on data the model has never encountered. This is because member samples have been well memorized by the model; even after the unlearning process, the model's predictions on these samples remain stable, resulting in lower absolute impact scores. In contrast, for unseen data, the removal of part of the training set leads to the loss of certain feature representations, weakening the model’s generalization ability and causing greater shifts in its predictions on these samples.}
This asymmetry in influence leads to a statistical divergence in the model behavior shifts between members and non-members, which ultimately amplifies the membership privacy risk.

\section{Dual-View Inference Attack}
This section introduces \textbf{DVIA} (Dual-View Inference Attack), the attack that exploits the amplified privacy risks enabled by access to both the original and unlearned models. DVIA comprises two components: the {Unlearning Confidence Difference} (UCD) as the attack metric, and a likelihood ratio inference module for membership prediction.

\subsection{Unlearning Confidence Difference}
\begin{figure}[t!]
    \centering
    \begin{subfigure}[t]{0.66\linewidth}
        \centering
        \includegraphics[width=\linewidth]{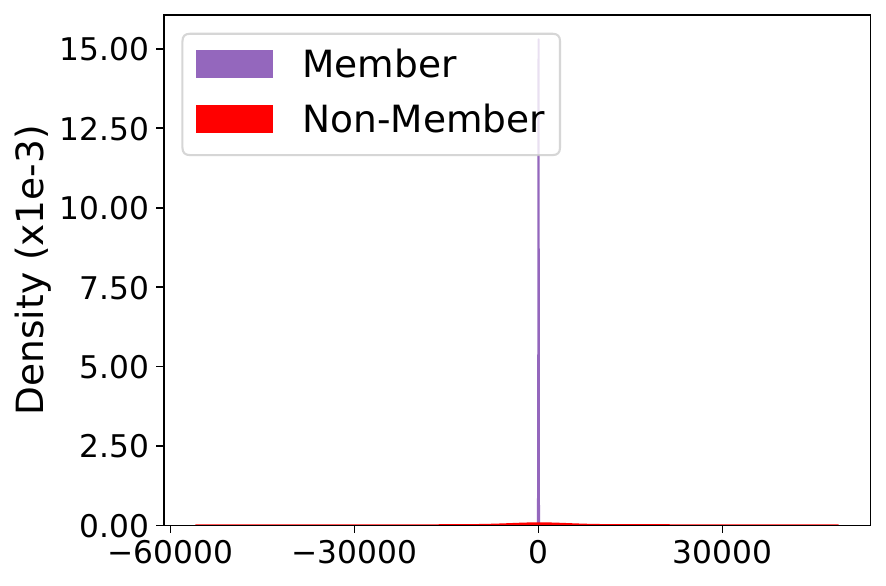}
        \caption{Influence Score}
    \end{subfigure}%
    
    \begin{subfigure}[t]{0.66\linewidth}
        \centering
        \includegraphics[width=\linewidth]{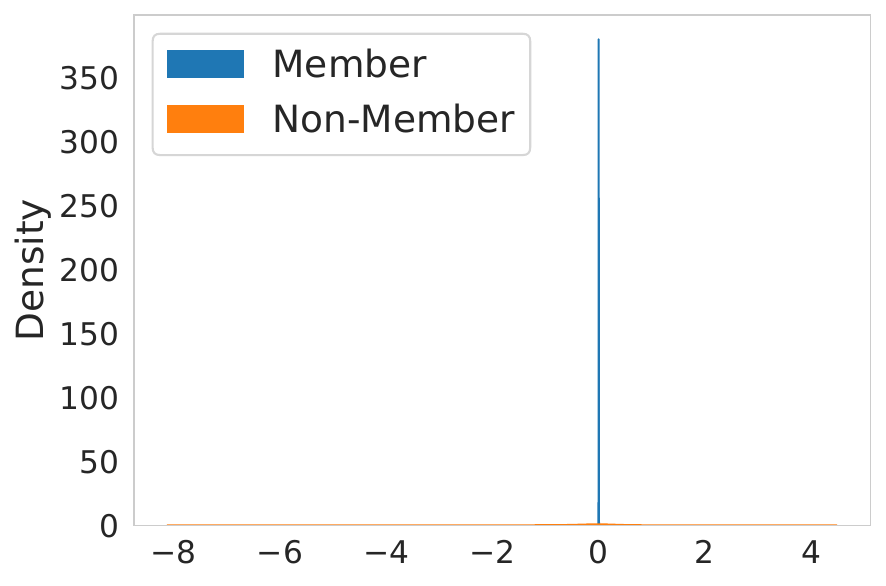}
        \caption{UCD}
    \end{subfigure}

    \caption{Density plots of influence score and UCD. It can be observed that the distributions of both show similarity.}
    \label{fig:Density}
\end{figure}

In this section, we design a behavioral impact–based attack metric under the DVIA framework. Our core insight is that machine unlearning induces notable differences in the model’s predictive behavior for member versus non-member data, as analyzed in Section~\ref{sec:gain}. This paper focuses on standard image classification tasks, where model behavior is primarily reflected in the confidence scores assigned to image labels. Therefore, we propose the Unlearning Confidence Difference (UCD) metric, which focuses on changes in the model's confidence when predicting the label of a given input. By leveraging dual queries to both the original and unlearned models, UCD reveals the privacy risks posed by machine unlearning on data that has not been forgotten. The details of the UCD metric are provided in Definition~\ref{def:ucd}.
\begin{definition}
\label{def:ucd}
The unlearning confidence difference (UCD) \(\Delta {(\mathbf{x}, y)}\) is defined as :
\begin{equation}
    \Delta{(\mathbf{x}, y)}= \theta^u(\mathbf{x})_y-\theta^o(\mathbf{x})_y, \nonumber
\end{equation}
where \(\theta^o\) represents the original trained model and \(\theta^u\) represents the unlearned model, \(\mathbf{\theta}(\mathbf{x})_y\) denotes the probability that the model \(\mathbf{\theta}\) maps the input \(\mathbf{x}\) to  correct label \(y\).
\end{definition}
Figure~\ref{fig:Density} illustrates the distribution of UCD scores for member and non-member data, as well as the impact of unlearning a specific data group on both. We observe a clear separation between the two distributions: after forgetting a group of data, the model retains high confidence in the samples it has memorized, while its confidence in previously unseen data fluctuates more noticeably, which aligns with our analysis in Section~\ref{sec:gain}.
Moreover, the behavioral patterns reflected by the UCD scores in the black-box setting closely mirror the influence scores computed in the white-box setting. This consistency demonstrates that the UCD metric effectively captures the differential impact of unlearning on member and non-member data, as shown in Figures~\ref{fig:inf}–\ref{fig:Density}.

\begin{algorithm}[t!]
\caption{Dual-View Inference Attack}
\begin{algorithmic}[1]
\STATE \textbf{Input:} Data distribution $\mathbb{D}$, target dataset $D_{\text{target}}$.
\STATE \textbf{Require:} Initialized model $\theta$, training algorithm $A$, unlearning algorithm $U$, number of shadow datasets $K$, decision threshold $\Lambda$.
\STATE ${S}_{\text{mem}} \gets \emptyset$ {\hfill $\rhd$ {Set to store UCD metrics for member}}
\STATE ${S}_{\text{non}} \gets \emptyset$ {\hfill $\rhd$ {Set to store UCD metrics for non-member}}
\FOR{$k = 1$ \textbf{to} $K$}
    \STATE Sample shadow datasets: $D_k^{\text{mem}}$, $D_k^{\text{non}}$ from $\mathbb{D}$
   \STATE Use $D_k^{\text{mem}}$, $D_k^{\text{non}}$, $A$ and $U$ to get the original model $\mathbf{\theta}_k^o$ and the unlearned model $\mathbf{\theta}_k^u$
    \STATE ${S}_{\text{mem}} \gets {S}_{\text{mem}} \cup \{\phi(\Delta(\mathbf{x},y)) \mid (\mathbf{x},y) \in D_k^{\text{mem}}\}$
    \STATE ${S}_{\text{non}} \gets {S}_{\text{non}} \cup \{\phi(\Delta(\mathbf{x},y)) \mid (\mathbf{x},y) \in D_k^{\text{non}}\}$
\ENDFOR
\STATE $\mu_{\text{mem}}, \sigma_{\text{mem}}^2 \gets \text{mean}({S}_{\text{mem}}), \text{var}({S}_{\text{mem}})$
\STATE $\mu_{\text{non}}, \sigma_{\text{non}}^2 \gets \text{mean}({S}_{\text{non}}), \text{var}({S}_{\text{non}})$
\STATE $S_t \gets \{\phi(\Delta(\mathbf{x},y)) \mid (\mathbf{x},y) \in D_{\text{target}}\}$
\STATE $L(\alpha) \gets \frac{p(\alpha \mid \mathcal{N}(\mu_{\text{mem}}, \sigma_{\text{mem}}^2))}{p(\alpha \mid \mathcal{N}(\mu_{\text{non}}, \sigma_{\text{non}}^2))}$ for each $\alpha \in S_t$
\STATE \textbf{Return:} $\{1 \text{ if } L(\alpha) > \Lambda \text{ else } 0 \mid \alpha \in S_t\}$
\end{algorithmic}
\label{alg:bui}
\end{algorithm}

\subsection{Likelihood Ratio Inference}
Figure~\ref{fig:Density} shows that there is a clear distinction between the UCD distributions of members and non-members, with member data being more likely to fall within the member distribution rather than the non-member distribution. Based on this, we design the likelihood ratio module to perform non-forgotten inference by leveraging this noticeable distribution difference.
Specifically, in a machine unlearning scenario, an attacker could use shadow models to obtain the distributions of member and non-member data, then calculate the likelihood ratios of target samples with respect to these distributions to determine their membership status \( b \), classifying them as either 0 or 1.

However, a key challenge lies in constructing the member and non-member distributions to support membership inference tasks. Prior research~\cite{lira} has shown that Gaussian fitting is a common and effective strategy, as it requires only four statistical parameters, namely the means and standard deviations of member and non-member data, to approximate the distributions.
However, the domain of the UCD distribution is $(-1, 1)$, which is inconsistent with the support of the Gaussian distribution. Directly fitting a Gaussian distribution in the original space leads to boundary truncation issues, making such an approach inappropriate.

To address this, we consider applying a monotonic transformation to the UCD metric to expand its domain. Although \cite{lira} has proposed $\phi_0(\Delta) = \frac{\Delta}{1 - \Delta}$ as a common transformation function, this function is not well-defined for negative values, limiting its applicability to the UCD metric.
Therefore, we propose the following transformation function:
\begin{equation}
\phi({\Delta}) = \log \left( \frac{1 + {\Delta}}{1 - {\Delta}} \right).
\label{Eq:phi}
\end{equation}
which is strictly monotonic on $x \in (-1, 1)$ and maps this interval to $(-\infty, \infty)$. This transformation not only aligns with the support of the Gaussian distribution but also maintains local linearity near $x = 0$, preserving the separability of the original UCD scores. Therefore, it provides a sound basis for subsequent Gaussian modeling, which underpins our attack design described in Algorithm~\ref{alg:bui}.
In a nutshell, the attacker first samples the shadow datasets from $\mathbb{D}$ and trains shadow original and shadow unlearned models from these datasets (Lines 6-7). The attacker then collects UCD from the two shadow models for both member and non-member data (Lines 8-9) and use \(\phi\) to fits them into two Gaussian distributions (Lines 11-12). For any targeted data point \((\mathbf{x}, y)\), the attacker calculates $\phi(\Delta(\mathbf{x}, y))$ (Line 13) and determines its likelihood ratio \(L(\alpha)\) relative to the member and non-member distributions (Line 14).We assume \( \Lambda = 1 \) by default, which is a reasonable choice as it implies that a sample is classified as a member if its membership probability exceeds its non-membership probability.

\noindent \textbf{Advantages.}
Compared to the inference approaches based on attack models~\cite{MIA1,U_MIA}, our module does not require the selection and training of a dedicated attack model. Furthermore, unlike threshold-based methods~\cite{u-label}, our module avoids the complexity of carefully tuning an appropriate threshold; a default threshold of 1 is sufficient for effective inference.

\noindent\textbf{Comparison with~\cite{U_LIRA}.}
While both our method and U-LIRA employ likelihood ratio-based inference to perform membership inference in the unlearning setting, they differ fundamentally in focus and granularity. U-LIRA targets forgotten data using a confidence-based metric and performs inference at the single-sample level, requiring multiple shadow models to estimate membership distributions per target instance. In contrast, DVIA is the first to investigate the privacy risks for non-forgotten data. We introduce the UCD metric and focus on group-level separability between members and non-members by leveraging statistical patterns across multiple samples. This design enables efficient inference using only a few shadow models and reveals the systemic privacy leakage introduced by unlearning.

\section{Experiment}
\subsection{Experimental Setup}
We conduct our experiments in Python 3.8 using PyTorch~\cite{pytorch}. Our evaluations are performed on two commonly used datasets, CIFAR10~\cite{cifar} and SVHN~\cite{svhn}. We divide the datasets into four parts to create the training set, validation set, shadow training set, and shadow validation set, with proportions of 50\%, 10\%, 20\%, and 20\%, respectively. For the models, we use ResNet18~\cite{resnet18} and the convolutional neural network ALLCNN~\cite{ALLCNN}, both of which are widely adopted in unlearning tasks~\cite{IU2,sparse,Salun,IU3}. Following \cite{U_MIA}, we train 4 shadow original models and 16 shadow unlearned models for each shadow original model. 
We assume that the attacker invokes the unlearning APIs to request the removal of 5\% of the training data. We use the best exact unlearning method, Retrain, as the default unlearning approach, and explore approximate unlearning methods in Section~\ref{sec:app}.
Following~\cite{MIA1,U_LIRA}, we use balanced accuracy as the primary evaluation metric, as it effectively measures the overall performance of the inference attack. We construct an evaluation dataset of 1,000 points, with half randomly sampled from the non-forgotten set and the other half from the validation set. 

\subsection{Comparative Experiment}
\begin{figure}[b!]
    \centering
    \begin{subfigure}[t]{0.48\linewidth}
        \centering
        \includegraphics[width=\linewidth]{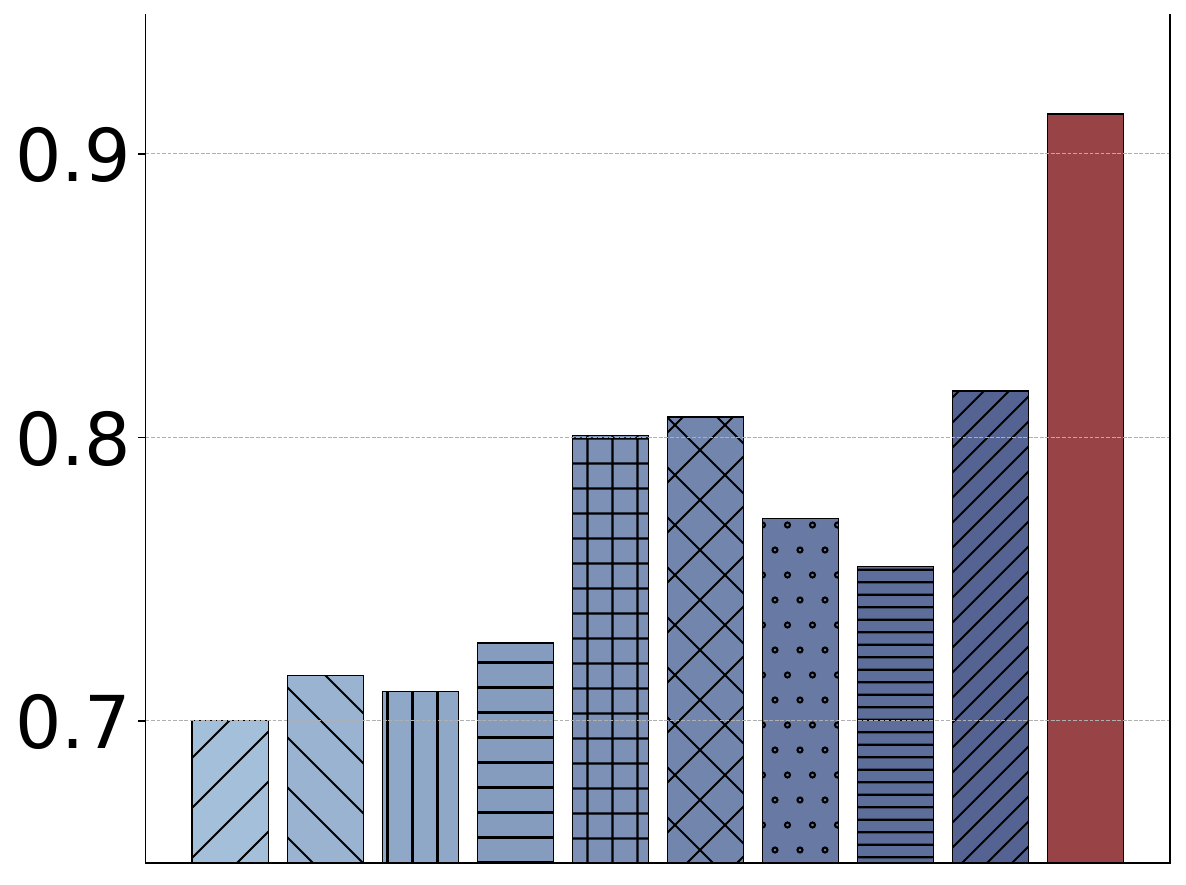}
        \caption{CIFAR10, ALLCNN}
    \end{subfigure}%
    \hfill
    \begin{subfigure}[t]{0.48\linewidth}
        \centering
        \includegraphics[width=\linewidth]{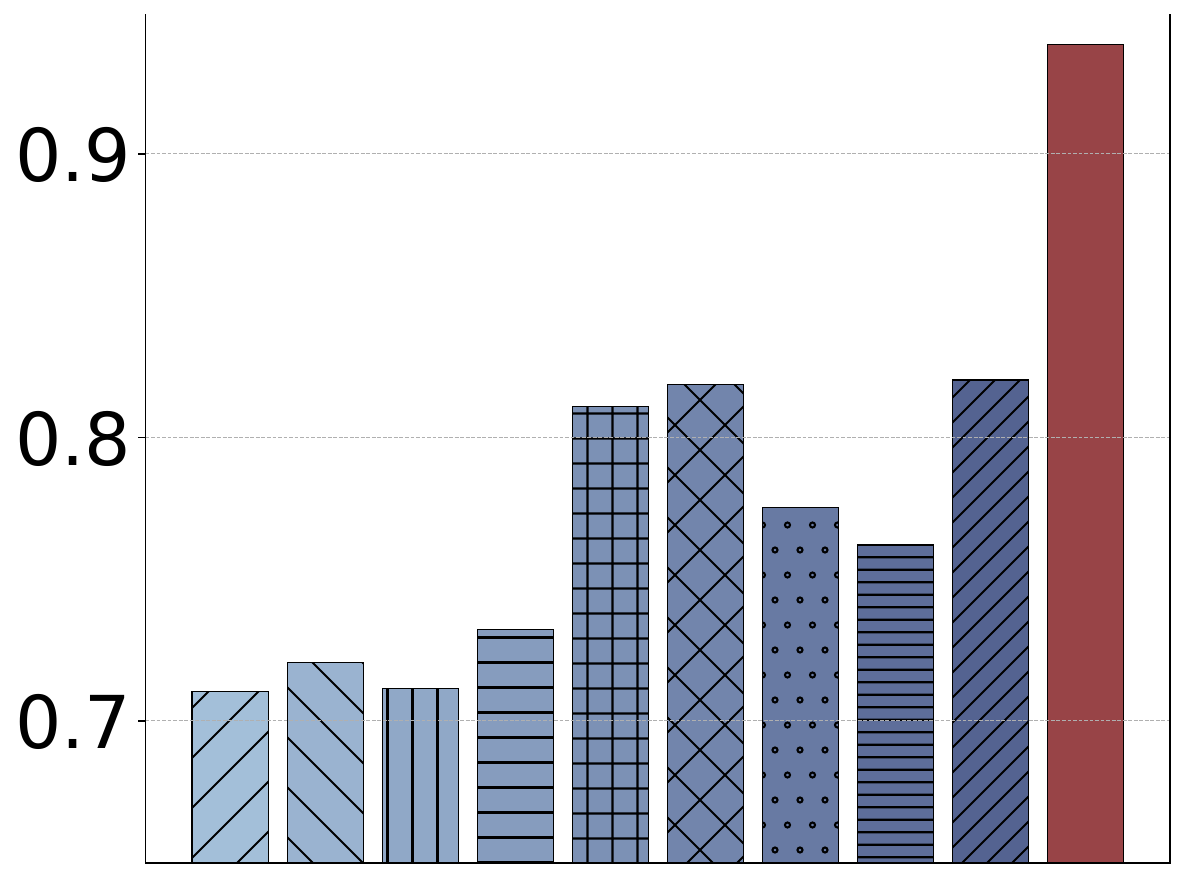}
        \caption{CIFAR10, ResNet18}
    \end{subfigure}
    \begin{subfigure}[t]{0.48\linewidth}
        \centering
        \includegraphics[width=\linewidth]{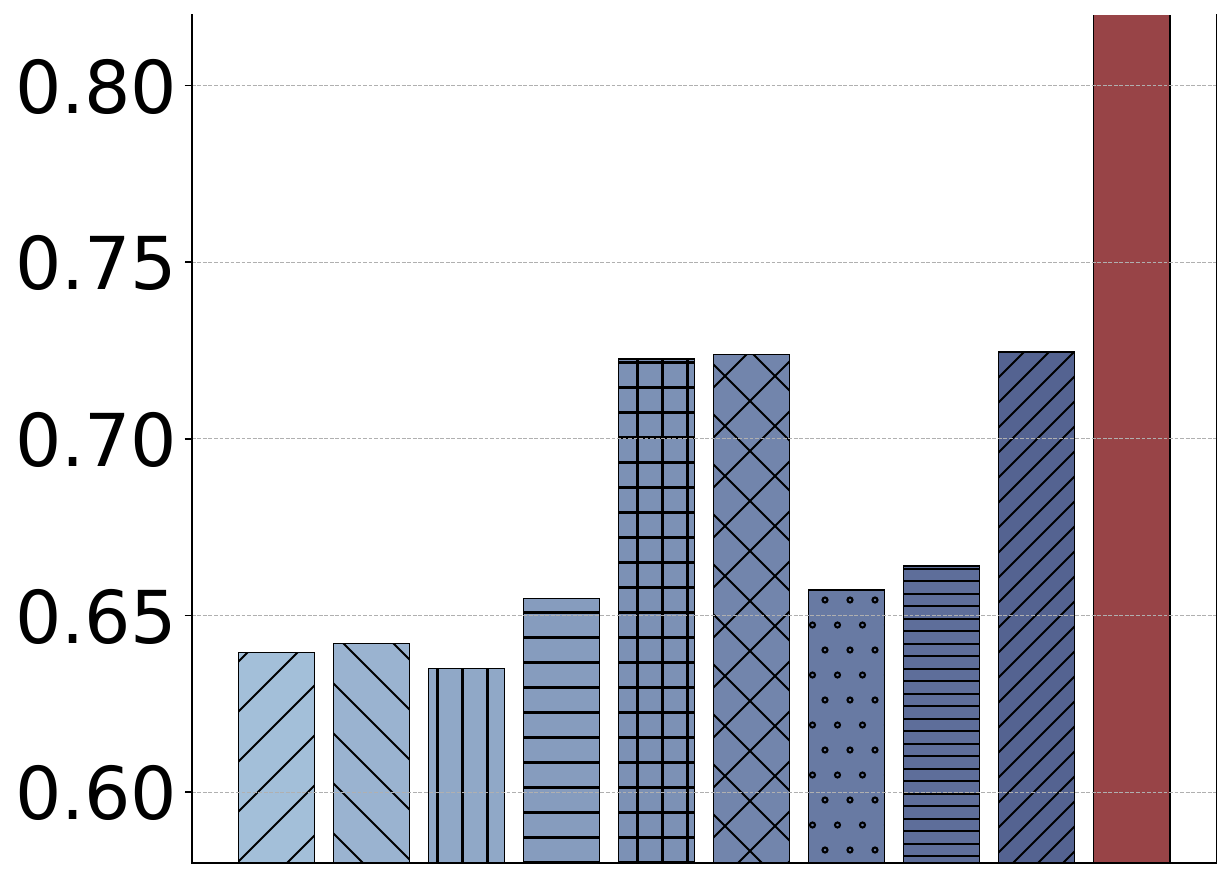}
        \caption{SVHN, ALLCNN}
    \end{subfigure}%
    \hfill
    \begin{subfigure}[t]{0.48\linewidth}
        \centering
        \includegraphics[width=\linewidth]{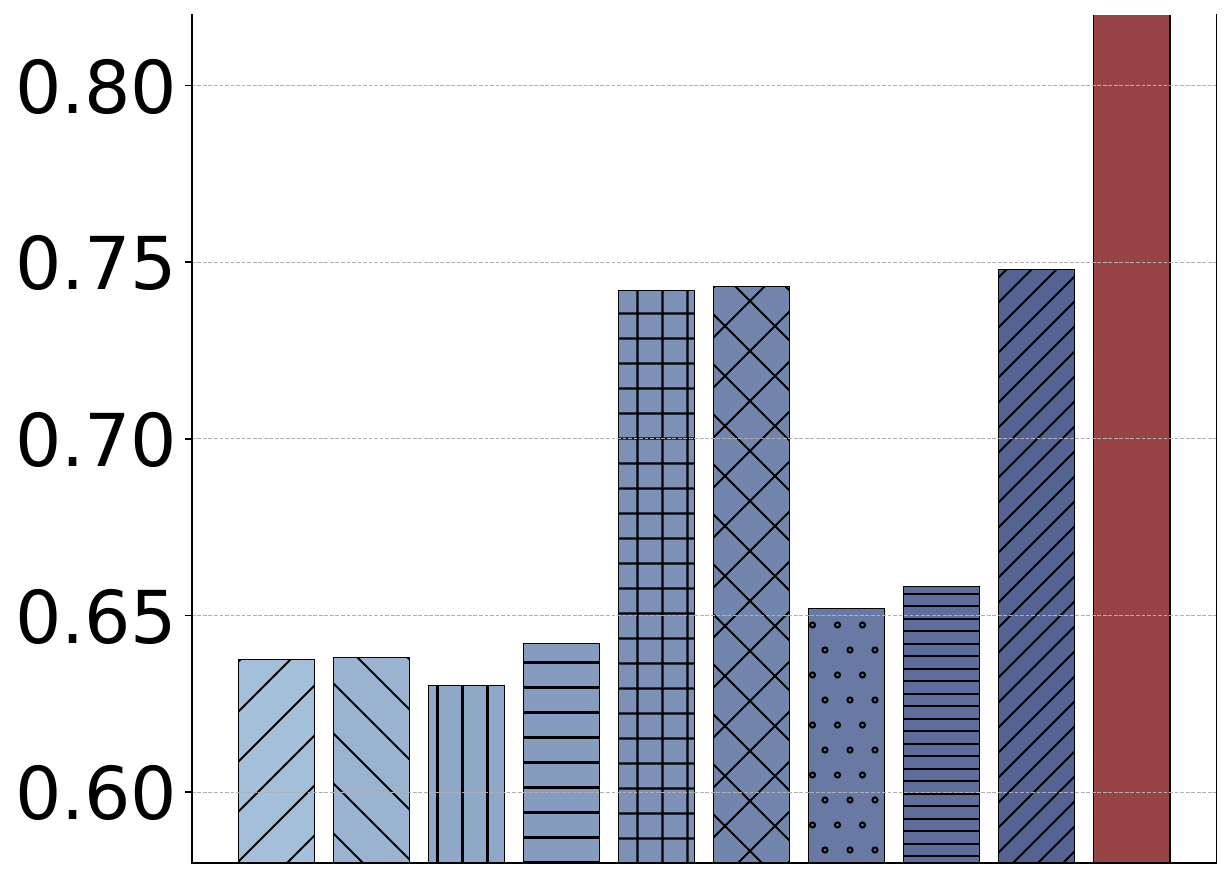}
        \caption{SVHN, ResNet18}
    \end{subfigure}
      \begin{subfigure}[t]{0.82\linewidth}
        \centering
        \includegraphics[width=\linewidth]{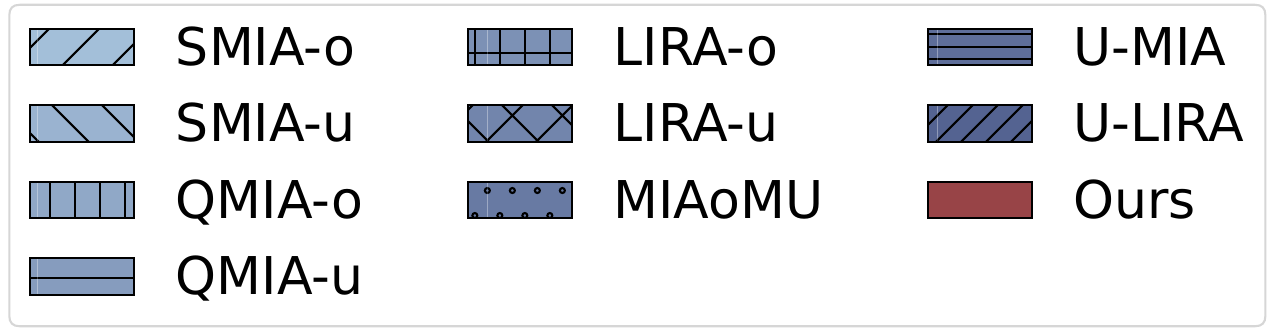}
        \caption{Legend illustration for different methods}
    \end{subfigure}
    \caption{Evaluation across different datasets and models, where the blue bars represent the comparative methods and the red bars represent our approach.}
    \label{fig:acc_comp}
\end{figure}

Our comparative study includes the following state-of-the-art membership inference attacks: LIRA~\cite{lira}, SMIA~\cite{MIA1}, and QMIA~\cite{label_only}. We evaluate their performance on both the original and unlearned models, denoted with the suffixes “-o” (original) and “-u” (unlearned), respectively. In addition, we consider advanced attacks that were originally designed to infer the membership of forgotten data, including MIAoMU~\cite{u-label}, U-LIRA~\cite{U_LIRA}, and U-MIA~\cite{U_MIA}, and apply them to non-forgetting inference scenario. We use these methods with their default settings.
The results are shown in Figure~\ref{fig:acc_comp}. It can be observed that LIRA, SMIA, and QMIA perform slightly better on the unlearned model than on the original model, and our method significantly outperforms all existing methods.

\begin{figure*}[t!]
    \begin{subfigure}[t]{0.47\linewidth}
        \centering
        \includegraphics[width=\linewidth]{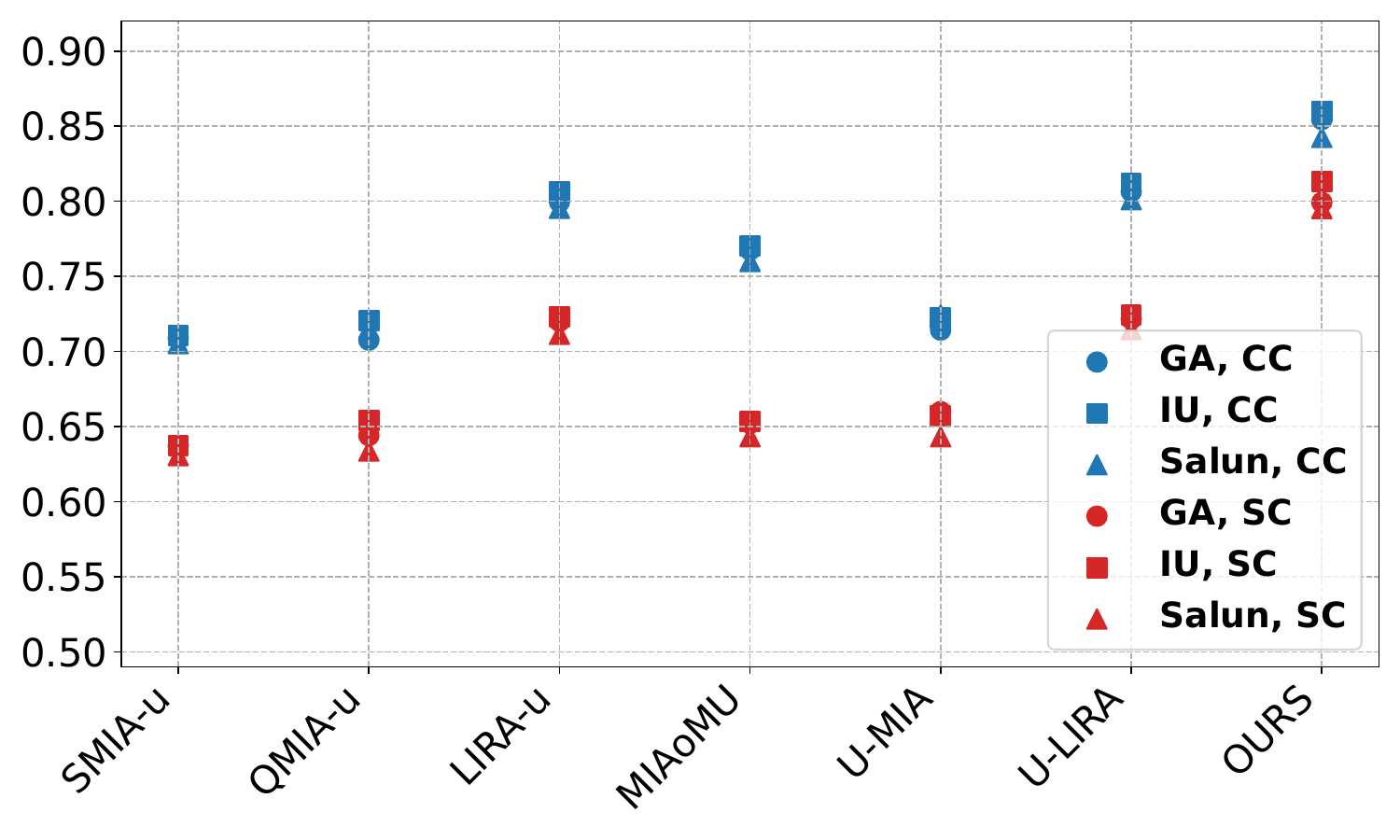}
        \caption{Attack with ALLCNN}
    \end{subfigure}%
    \begin{subfigure}[t]{0.47\linewidth}
        \centering
        \includegraphics[width=\linewidth]{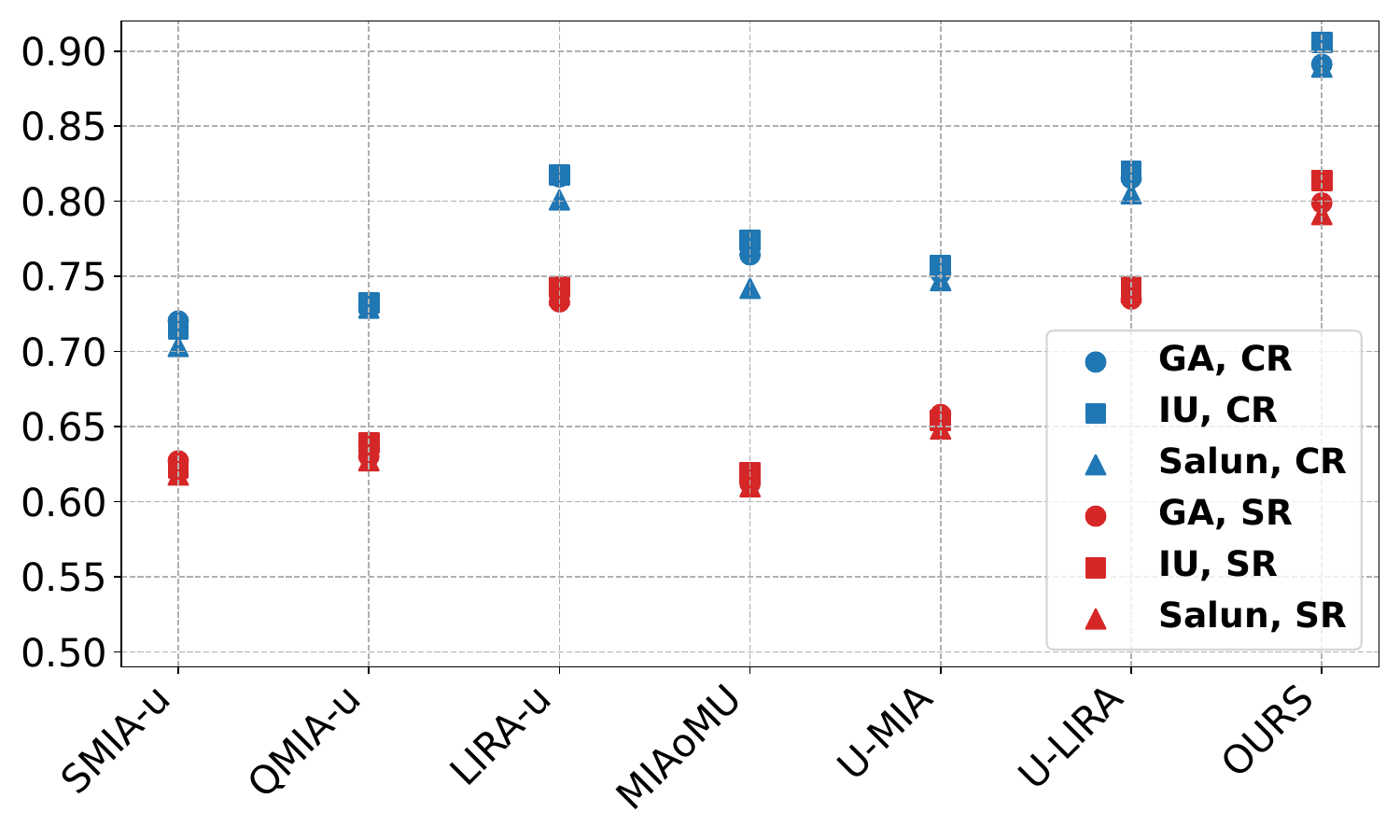}
        \caption{Attack with ResNet18}
    \end{subfigure}

    \caption{The evaluation of attack performance under different unlearning methods.}
    \label{fig:app}
\end{figure*}

\subsection{Ablation Experiments}

\textbf{Different Attack Modules.}
We replace the likelihood ratio inference (LRI) module with a decision tree (DT)~\cite{DT} and a threshold method~\cite{label_only}. As shown in Table~\ref{tab:da}, LRI achieves superior attack performance. This is because likelihood ratios typically capture distributional differences more effectively, leading to more accurate membership inference.

\noindent \textbf{Different Attack Metrics.} We evaluate the attack performance of the UCD metric, original model confidence $\text{Conf}_o$, and unlearned model confidence $\text{Conf}_u$, with the attack model using a likelihood ratio inference module. The results are shown in Table~\ref{tab:attack_metrics}. It is evident that the inference attack based on the UCD metric performs the best, further highlighting the privacy risks posed by unlearning APIs.
\begin{table}[t]
  \centering
   \scalebox{0.8}{
    \begin{tabular}{lrrrr}
    \toprule
    \multirow{2}[4]{*}{} & \multicolumn{2}{c}{CIFAR10} & \multicolumn{2}{c}{SVHN} \\
 \cmidrule(r){2-3}\cmidrule(r){4-5}          & \multicolumn{1}{l}{ALLCNN} & \multicolumn{1}{l}{ResNet18} & \multicolumn{1}{l}{ALLCNN} & \multicolumn{1}{l}{ResNet18} \\
  \cmidrule(r){1-3}\cmidrule(r){4-5}
    LRI   & 0.9141 & 0.9386 & 0.8222 & 0.8210 \\
    DT   & 0.8665 & 0.8525 & 0.7321 & 0.7601 \\
    Threshold  & 0.8612 & 0.875 & 0.7422 & 0.7602 \\
    \bottomrule
    \end{tabular}%
    }
    \caption{Evaluation with different attack modules.}

  \label{tab:da}%
\end{table}%

\begin{table}[tb!]
  \centering
  \scalebox{0.8}{
\begin{tabular}{lrrrr}
    \toprule
    \multirow{2}[4]{*}{} & \multicolumn{2}{c}{CIFAR10} & \multicolumn{2}{c}{SVHN} \\
 \cmidrule(r){2-3}\cmidrule(r){4-5}         & \multicolumn{1}{l}{ALLCNN} & \multicolumn{1}{l}{ResNet18} & \multicolumn{1}{l}{ALLCNN} & \multicolumn{1}{l}{ResNet18} \\
   \cmidrule(r){1-3}\cmidrule(r){4-5}
     UCD   & 0.9141  & 0.9386  & 0.8222  & 0.8210  \\
    $\text{Conf}_o$ & 0.8221  & 0.8420  & 0.6874  & 0.6995  \\
    $\text{Conf}_u$ & 0.8330  & 0.8575  & 0.6960  & 0.7018  \\
    \bottomrule
    \end{tabular}%
    }
      \caption{Evaluation of attack metrics.}
  \label{tab:attack_metrics}%
 
\end{table}%

\subsection{Approximate Unlearning}
\label{sec:app}
We evaluate our method with approximate unlearning from the perspectives of comp. overhead and attack effect. 
We select several advanced approximate unlearning methods and apply them to non-forgotten inference, including Gradient Ascent (GA)~\cite{IU2,GA1,GA3}, Influence Unlearning (IU)~\cite{sparse,Salun} and Salun~\cite{Salun}. 

\noindent\textbf{Comp. Overhead.} Figure~\ref{fig:more_result} (a) presents the comp. overhead of training unlearned shadow models using different unlearning methods. It is evident that approximate unlearning is significantly more efficient than exact unlearning.

\noindent\textbf{Attack Evaluation}. Figures \ref{fig:acc_comp} and \ref{fig:app} indicate that membership inference attacks are generally less effective under approximate unlearning compared to retraining unlearning. This is because approximate unlearning often has a greater impact on non-forgotten data~\cite{GA1}.
Although DVIA, based on approximate unlearning, is less effective than exact unlearning methods in attack performance, it still outperforms comparative membership inference methods, highlighting the elevated privacy risks in the dual-view setting.  In future work, we aim to further explore the trade-off between attack effectiveness and comp. efficiency in non-forgotten inference.


\subsection{Further Discussions}
\label{sec:discuss}

\noindent \textbf{Impact of Unlearning Ratio Knowledge.}
As shown in Table~\ref{tab:k}, the attack remains effective even when the attacker does not know the exact unlearning ratio. This is because machine unlearning significantly affects non-members while having minimal impact on members, and this trend holds robustly across different unlearning ratio settings.

\begin{table}[t]
  \centering
   \scalebox{0.8}{
    \begin{tabular}{lrrrr}
    \toprule
         & \multicolumn{1}{l}{CC} & \multicolumn{1}{l}{CR} & \multicolumn{1}{l}{SC} & \multicolumn{1}{l}{SR} \\
    \midrule
    (5\%,5\%) & 0.9141  & 0.9386  & 0.8222  & 0.8210  \\
    (5\%,10\%) & 0.9090  & 0.9372  & 0.8134  & 0.7956  \\
    (10\%,5\%) & 0.9128  & 0.9344  & 0.8075  & 0.8165  \\
    (10\%,10\%) & 0.9220  & 0.9425  & 0.8212  & 0.8182  \\
    \bottomrule
    \end{tabular}%
    }
   \caption{Impact of unlearning proportions in target and shadow models. Each \((a\%, b\%)\) pair denotes the unlearning ratios of the target and shadow models, respectively.}
  \label{tab:k}%
\end{table}%

\noindent\textbf{Defense.} 
\noindent We implement differential privacy~\cite{dp1,dp2} by using Opacus~\cite{opacus} and setting the noise\_multiplier to 1.0. The experimental results are shown in Table~\ref{tab:dp}. Although differential privacy can defend against our attacks, it may make the model unusable.
\begin{table}[t]
  \centering
  \scalebox{0.8}{
    \begin{tabular}{lrrrrrr}
      \toprule
      & {Drop (\%)} & {Retrain} & {GA} & {IU} & {Salun} \\
      \midrule
      {CC} & 32.64 & 0.5248 & 0.5135 & 0.5293 & 0.5250 \\
      {CR} & 33.14 & 0.5293 & 0.5351 & 0.4925 & 0.5162 \\
      {SC} & 42.81 & 0.5128 & 0.4919 & 0.5048 & 0.5043 \\
      {SR} & 39.73 & 0.5151 & 0.5103 & 0.5103 & 0.5120 \\
      \bottomrule
    \end{tabular}
  }  
  \caption{Evaluation under differential privacy. {Drop} indicates the drop in model accuracy on the validation set.}
  \label{tab:dp}
\end{table}

\begin{figure}[t!]
    \centering
    \begin{subfigure}[t]{0.48\linewidth}
        \centering
        \includegraphics[width=\linewidth]{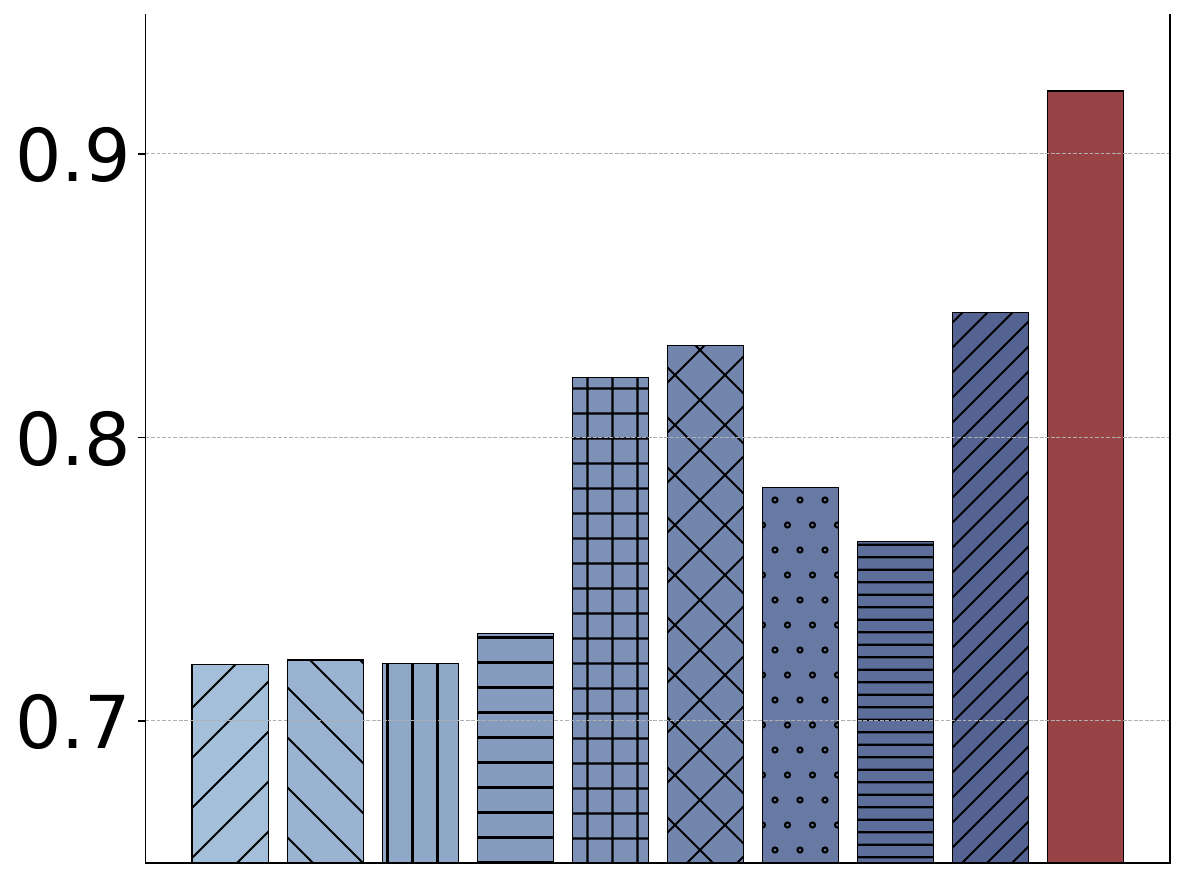}
        \caption{CIFAR10, ALLCNN}
    \end{subfigure}%
    \hfill
    \begin{subfigure}[t]{0.48\linewidth}
        \centering
        \includegraphics[width=\linewidth]{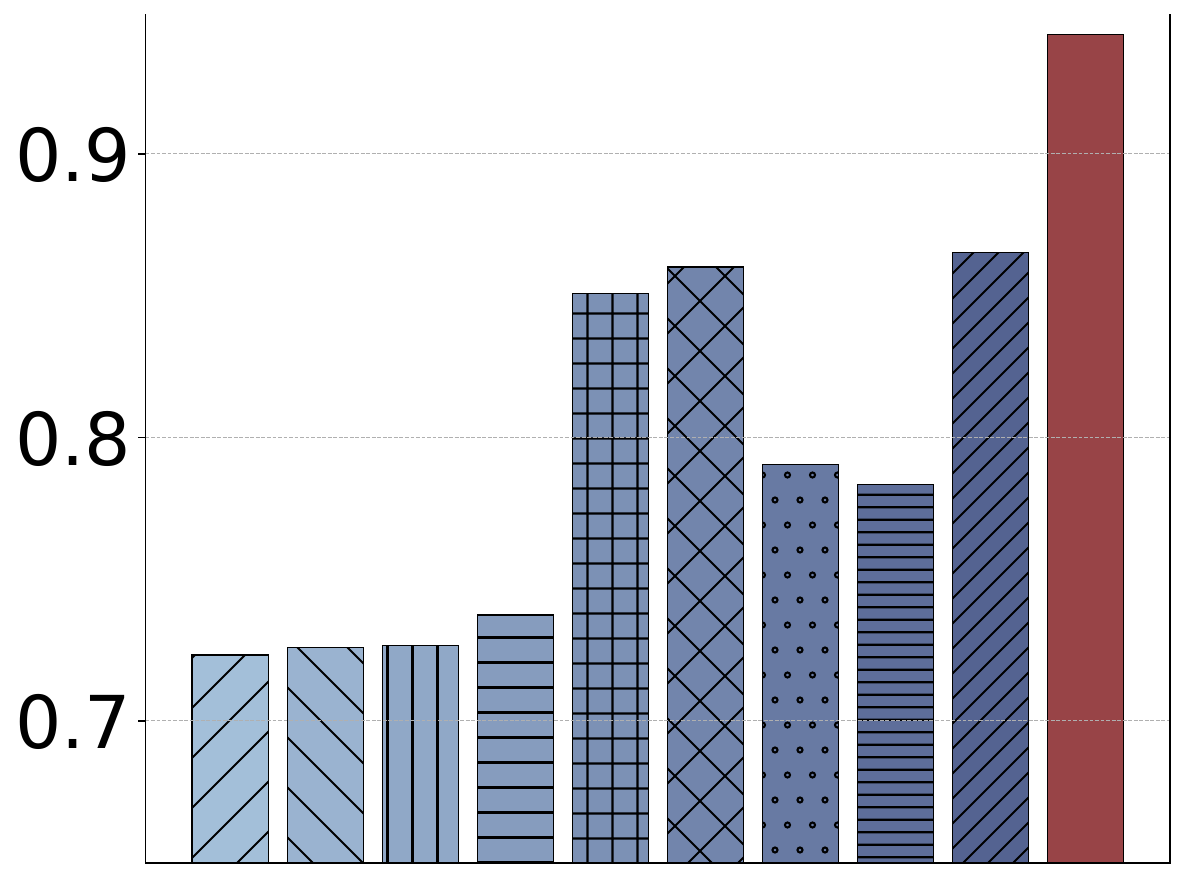}
        \caption{CIFAR10, ResNet18}
    \end{subfigure}
    
    \begin{subfigure}[t]{0.48\linewidth}
        \centering
        \includegraphics[width=\linewidth]{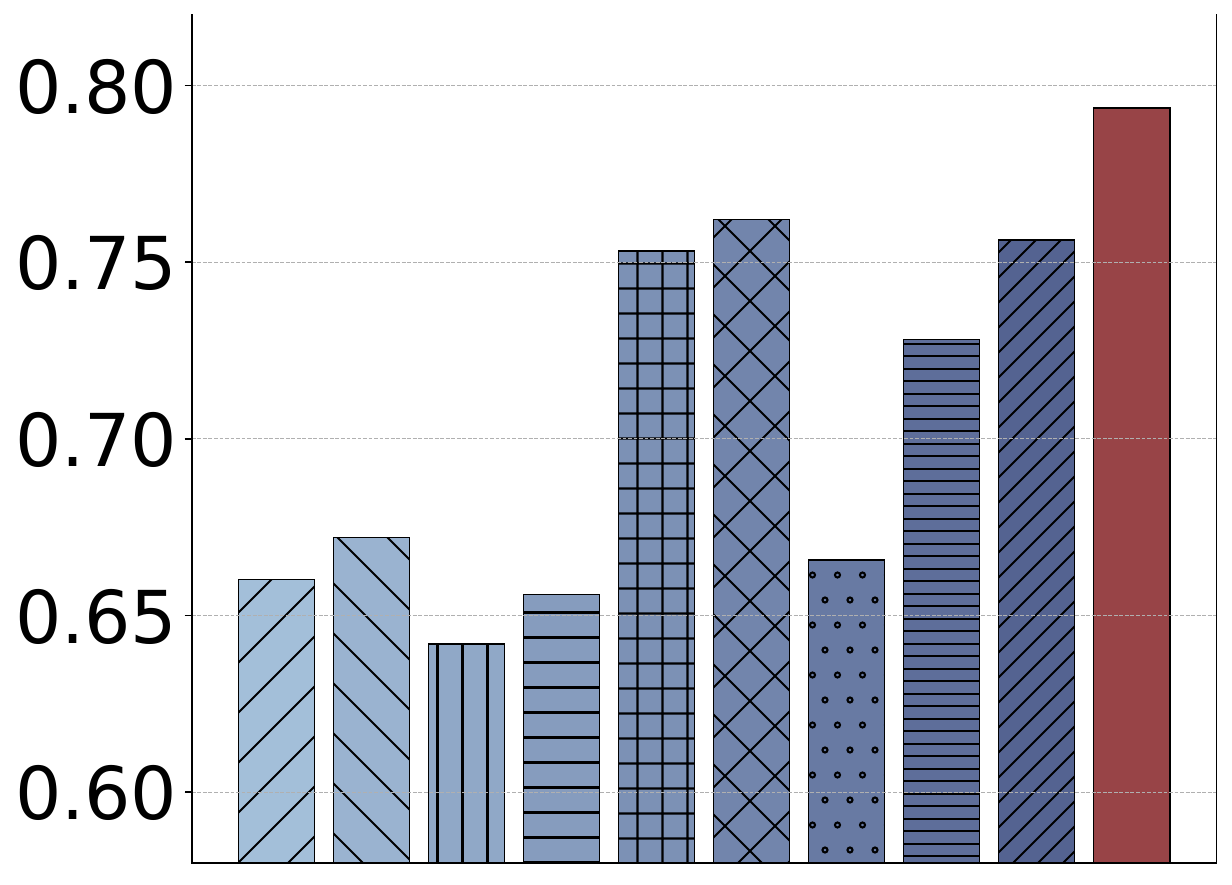}
        \caption{SVHN, ALLCNN}
    \end{subfigure}%
    \hfill
    \begin{subfigure}[t]{0.48\linewidth}
        \centering
        \includegraphics[width=\linewidth]{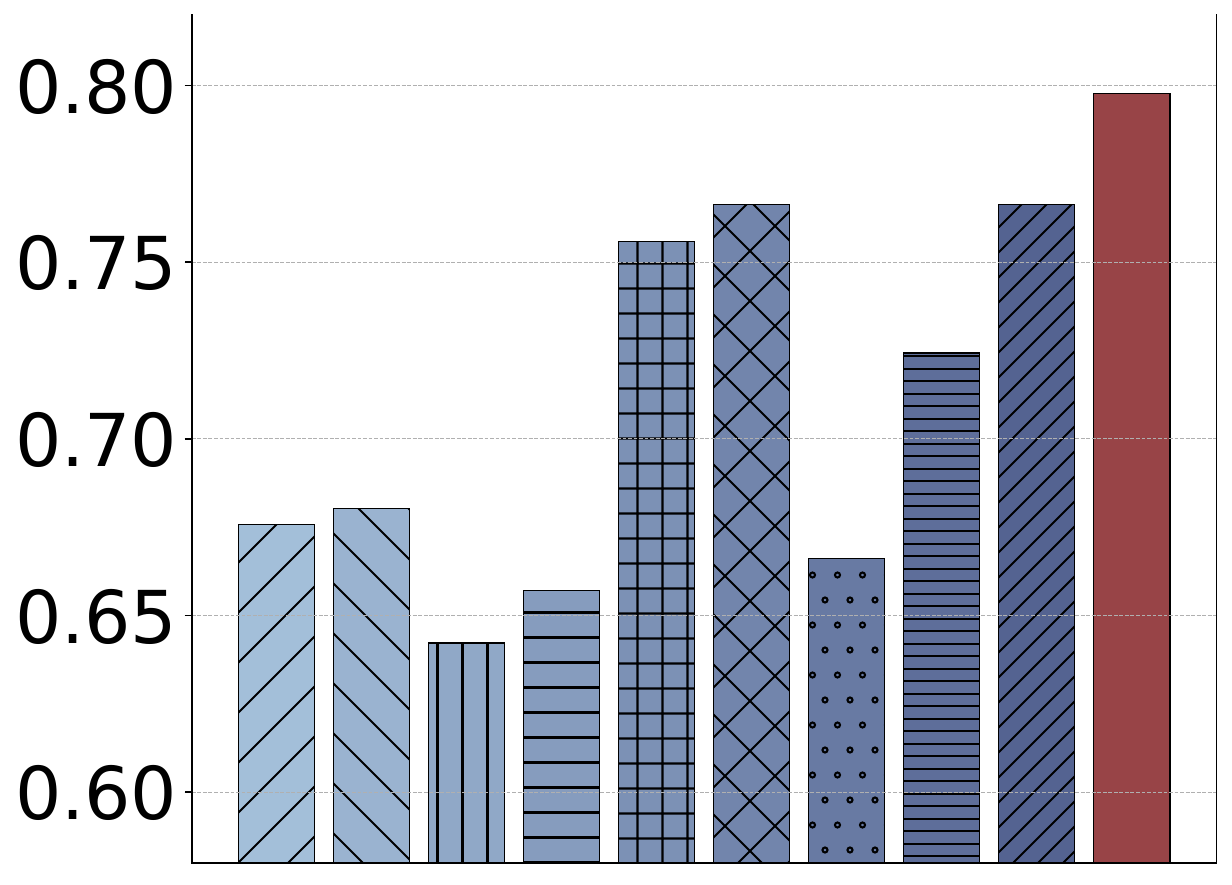}
        \caption{SVHN, ResNet18}
    \end{subfigure}
\begin{subfigure}[t]{0.82\linewidth}
        \centering
        \includegraphics[width=\linewidth]{fig/method_legend.pdf}
        \caption{Legend illustration for different methods}
    \end{subfigure}
    \caption{The AUC evaluation across different methods.}
    \label{fig:auc_comp}
       
\end{figure}

    

\begin{figure}[t!]
    \centering
    \begin{subfigure}[t]{0.49\linewidth}
        \centering
        \includegraphics[width=\linewidth]{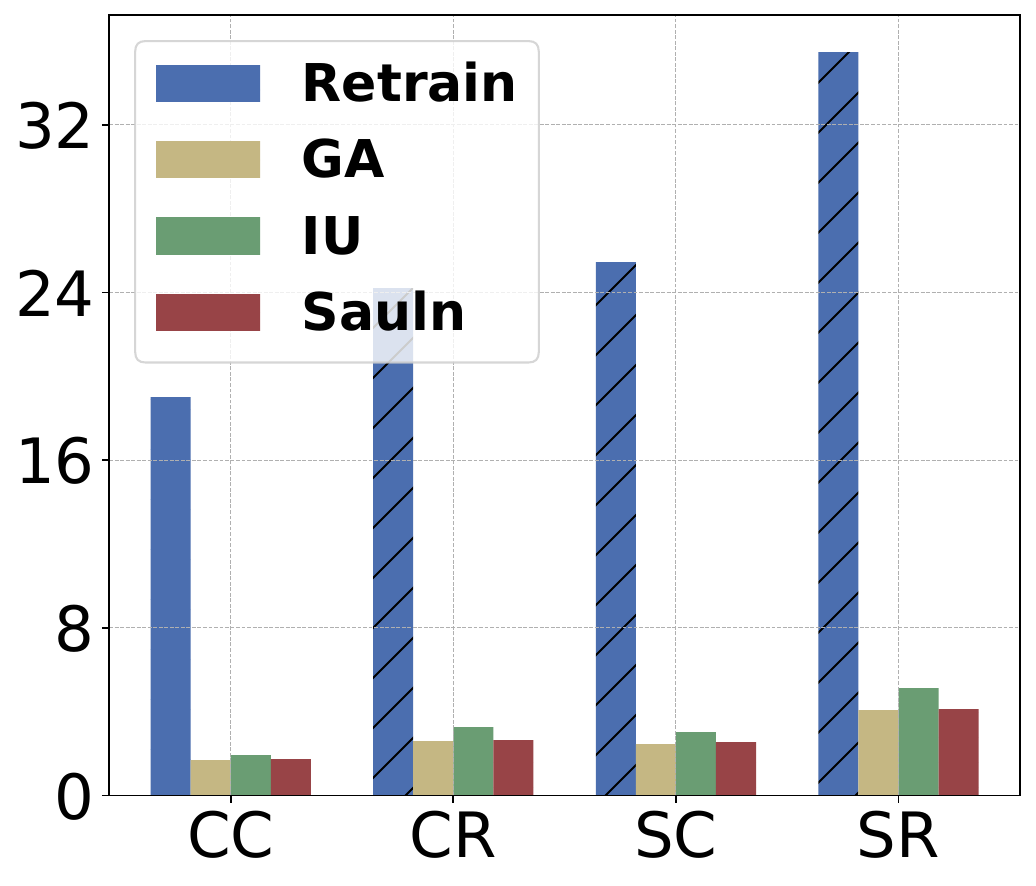}
        \caption{Comp. Overheads}
    \end{subfigure}%
    \begin{subfigure}[t]{0.49\linewidth}
        \centering
        \includegraphics[width=\linewidth]{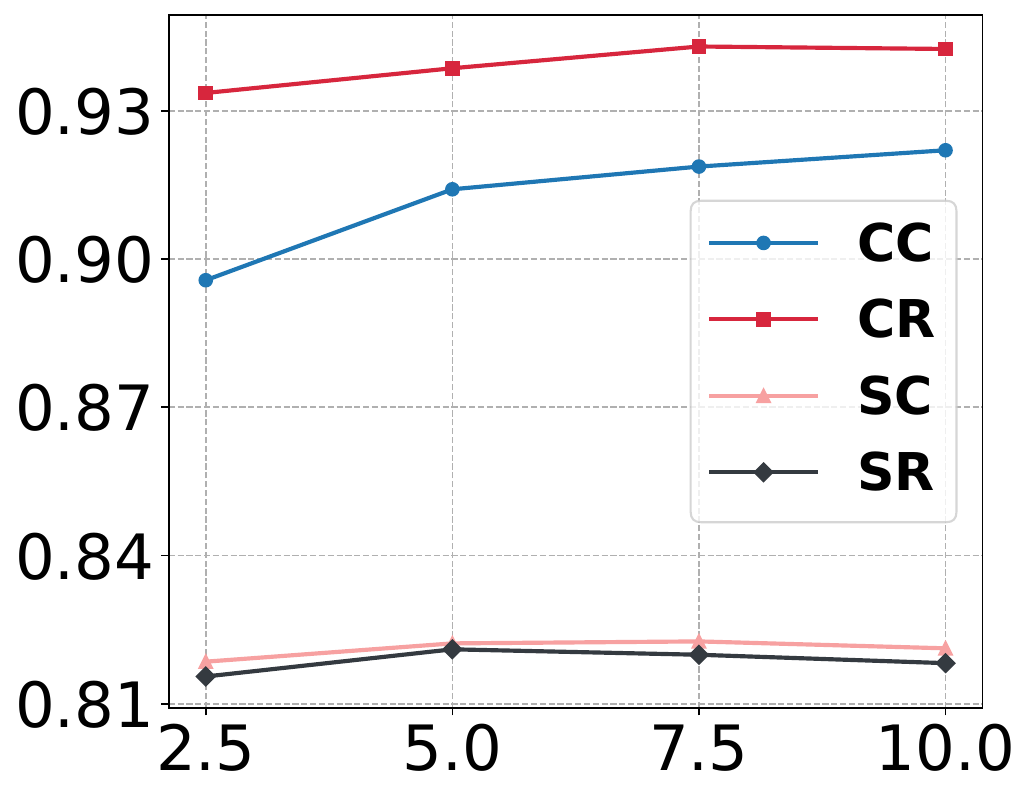}
        \caption{Different unlearning ratios.}
    \end{subfigure}%
    \caption{More evaluation results.}
    \label{fig:more_result}
\end{figure}


    

\noindent \textbf{More Metrics.} We present additional comparisons using AUC. As shown in Figure~\ref{fig:auc_comp}, our method outperforms others.  Additional metrics can be found in the appendix.

\noindent \textbf{Impact of Different Unlearning Ratios.}
Figure~\ref{fig:more_result} (b) illustrates the performance of DVIA under varying unlearning ratios. Notably, even with as little as 2.5\% of data unlearned, DVIA achieves strong performance.

\noindent \textbf{Evaluation on CIFAR100~\cite{cifar}.} Following \cite{lira}, we evaluate DVIA on CIFAR100, where it achieves 0.8975 accuracy on AllCNN and 0.9444 on ResNet18, demonstrating strong performance.

\section{Conclusion, Limitation, and Future}\label{conclusion}
In this paper, we challenge the notion that machine unlearning ensures privacy and, for the first time, reveal the privacy risks it poses to non-forgotten data through an information-theoretic lens. We introduce the novel {dual-view} setting, where adversaries can query both the original and the unlearned models, and demonstrate that such dual access significantly amplifies privacy leakage. To formalize this advantage, we propose the notion of {privacy information gain}, which highlights the increased privacy exposure resulting from comparing the outputs of the two models. To exploit this vulnerability in practice, we propose DVIA,  which uses black-box queries and a lightweight likelihood ratio module to infer membership of retained data. Our experiments across diverse datasets and model architectures validate the effectiveness of DVIA and underscore the privacy vulnerabilities introduced by unlearning in dual-view scenarios.

In terms of limitations, we focus solely on image classification tasks, which are the primary application area for machine unlearning. In future work, we plan to broaden our analysis to other domains such as natural language processing and graph learning. Furthermore, we intend to investigate the trade-offs between inference accuracy, computational efficiency, and privacy guarantees in approximate unlearning frameworks.



\section*{Acknowledgments}
Minghui’s work is supported in part by the National Natural Science Foundation of China (Grant No.62572206).

\section{Theoretical Proof.}
\begin{theorem}
For the target dataset \( D_t \), The privacy knowledge gain satisfies:
\begin{align*}
\textnormal{Gain}(D_t) > 0 
\quad &\text{if and only if} \\
\mathbb{P}(\delta(D_t) \mid M_{D_t} = 1) 
&\ne 
\mathbb{P}(\delta(D_t) \mid M_{D_t} = 0),
\end{align*}
where \( \delta(D_t)\) is the behavioral impact for \( D_t \),
\( M_{D_t} \) is the membership status of \( D_t \), and \( \mathbb{P}(\cdot \mid \cdot) \) denotes conditional probability.
\label{theo:t1}
\end{theorem}
\begin{proof}
We prove both directions.

 If \( \text{Gain}(D_t) > 0 \), then \( \mathbb{P}(\delta(D_t) \mid M=1) \ne \mathbb{P}(\delta(D_t) \mid M=0) \):

By definition,
\[
\begin{aligned}
\text{Gain}(D_t) 
&= I(M_{D_t}; s_o(D_t), s_u(D_t)) - I(M_{D_t}; s(D_t)), \\
&= I(M; Z) - I(M; s(D_t)),
\end{aligned}
\]
where \( Z = (s_o(D_t), s_u(D_t)) \) denotes the full dual-view behavior, and \( s(D_t) \in \{s_o(D_t), s_u(D_t)\} \) is a single-view observation.

If \( \text{Gain}(D_t) > 0 \), then \( I(M; Z) > I(M; s(D_t)) \), i.e., access to the full dual-view \( Z \) reveals more information about membership than the single-view.

Since \( \delta(D_t) = 1 - \text{sim}(s_o(D_t), s_u(D_t)) \) is a deterministic function of \( Z \), we apply the chain rule of mutual information:
\[
I(M; Z) \geq I(M; s(D_t)) + I(M; \delta(D_t) \mid s(D_t)).
\]

Therefore, \( \text{Gain}(D_t) = I(M; Z) - I(M; s(D_t)) \geq I(M; \delta(D_t) \mid s(D_t)) \). If \( \text{Gain}(D_t) > 0 \), then \( I(M; \delta(D_t) \mid s(D_t)) > 0 \), which implies that \( \delta(D_t) \) carries additional information about \( M \) not present in \( s(D_t) \).

In particular, this requires the conditional distributions to differ:
\[
\mathbb{P}(\delta(D_t) \mid M=1) \ne \mathbb{P}(\delta(D_t) \mid M=0).
\]

If \( \mathbb{P}(\delta(D_t) \mid M=1) \ne \mathbb{P}(\delta(D_t) \mid M=0) \), then \( \text{Gain}(D_t) > 0 \):

This implies \( I(M; \delta(D_t)) > 0 \). Since \( \delta(D_t) \) is a deterministic function of \( Z \), and \( s(D_t) \) is only a partial view of \( Z \), we again apply the chain rule:
\[
I(M; Z) \geq I(M; s(D_t)) + I(M; \delta(D_t) \mid s(D_t)).
\]

Hence,
\[
\text{Gain}(D_t) = I(M; Z) - I(M; s(D_t)) \geq I(M; \delta(D_t) \mid s(D_t)).
\]

If \( I(M; \delta(D_t)) > 0 \), and this information is not fully contained in \( s(D_t) \), then \( I(M; \delta(D_t) \mid s(D_t)) > 0 \), leading to \( \text{Gain}(D_t) > 0 \).
\end{proof}

\section{Further Details and Discussion}
This section presents the related experimental setup and more experimental results.
\subsection{Experimental setup}
We conduct our experiments in PyTorch using four GeForce RTX 3090 GPUs. We train models with batchsize=64, learning rate lr=0.01. We use SGD optimizer with momentum of 0.9 and set epoch=200. We selected 5,000 data points separately from the retain set and the validation set to evaluate the model's performance.

\subsection{Unlearning Methods.}
We selected several state-of-the-art approximate unlearning methods to implement approximate non-forgotten inference, as detailed below.

\textbf{Gradient Ascent (GA)}~\cite{GA1,IU2}:GA erases the model’s memory of data points by performing gradient ascent on the data points to be forgotten.

\textbf{Influence Unlearning (IU)}~\cite{IU2,sparse,Salun}: IU uses influence functions to eliminate the impact of certain data points on model parameters. Following \cite{sparse,Salun}, we perform unlearning using an influence function based on Woodfisher~\cite{woodfisher}.

\textbf{Salun}~\cite{Salun}: 
Salun makes the model forget specific data points by adjusting certain weights in the model. 

\subsection{More Experimental Results}
\textbf{Model Performance.} We present the model performance under different unlearning methods in Table~\ref{tab:dff_model_perform}. It can be observed that approximate unlearning methods have a more significant impact on non-forgotten data compared to retraining. Additionally, we find that on CIFAR10, exact  unlearning affects ResNet18 more significantly than ALLCNN. This is because ResNet18 has a more complex structure, making its performance more susceptible to changes after unlearning many features. It is worth noting that, whether in the ResNet18 model, where unlearning has a significant impact on accuracy, or in the ALLCNN model, where the impact is minimal, our method achieves good results. This is because the UCD metric clearly distinguishes between members and non-members, enabling effective attacks even when unlearning causes only minor changes in the model's accuracy.

\begin{table}[t]
  \centering
 
    \scalebox{0.8}{
    \begin{tabular}{clrrrrr}
    \toprule
          &       & \multicolumn{1}{l}{Original} & \multicolumn{1}{l}{Retrain} & \multicolumn{1}{l}{GA} & \multicolumn{1}{l}{IU}  & \multicolumn{1}{l}{Salun} \\
    \midrule
    \multirow{3}[2]{*}{CC} & $D_{\text{retain}}$ & 100.00  & 100.00  & 97.92  & 98.64   & 98.24  \\
          & $D_{\text{val}}$  & 83.16  & 78.58  & 76.01  & 78.38   & 75.84  \\
          & $D_{\text{forget}}$ & 100.00  & 73.28  & 60.40  & 74.40  & 65.36  \\
    \midrule
    \multirow{3}[2]{*}{CR} & $D_{\text{retain}}$ & 100.00  & 100.00  & 98.41  & 98.86   & 98.60  \\
          & $D_{\text{val}}$  & 81.28  & 67.12  & 74.88  & 76.10  & 75.14  \\
          & $D_{\text{forget}}$ & 100.00  & 59.20  & 69.84  & 78.24   & 73.20  \\
    \midrule
    \multirow{3}[2]{*}{SC} & $D_{\text{retain}}$ & 100.00  & 100.00  & 99.73  & 98.74   & 99.43  \\
          & $D_{\text{val}}$  & 92.84  & 92.34  & 89.08  & 87.75   & 87.81  \\
          & $D_{\text{forget}}$ & 100.00  & 88.90  & 96.70  & 86.00   & 94.90  \\
    \midrule
    \multirow{3}[2]{*}{SR} & $D_{\text{retain}}$ & 100.00  & 100.00  & 99.56  & 99.63    & 99.61  \\
          & $D_{\text{val}}$  & 90.30  & 89.73  & 85.47  & 85.24   & 85.52  \\
          & $D_{\text{forget}}$ & 100.00  & 85.40  & 88.80  & 89.30    & 90.10  \\
    \bottomrule
    \end{tabular}%
    }
     \caption{Model Performance (\%) with Different Unlearning}
  \label{tab:dff_model_perform}%
\end{table}%

\begin{figure*}[t!]
    \centering
    \begin{subfigure}[t]{0.23\linewidth}
        \centering
        \includegraphics[width=\linewidth]{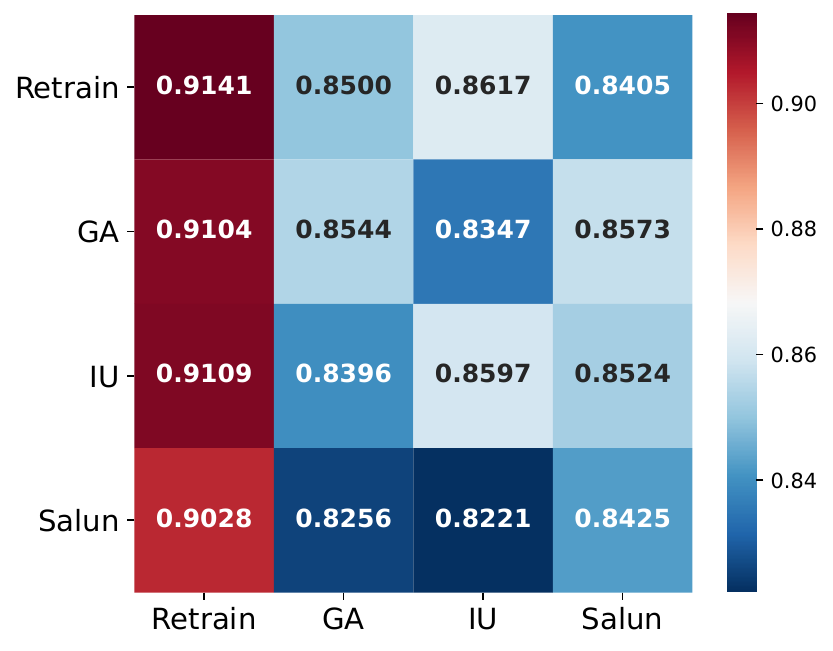}
        \caption{CIFAR10, ALLCNN}
    \end{subfigure}%
    \hfill
    \begin{subfigure}[t]{0.23\linewidth}
        \centering
        \includegraphics[width=\linewidth]{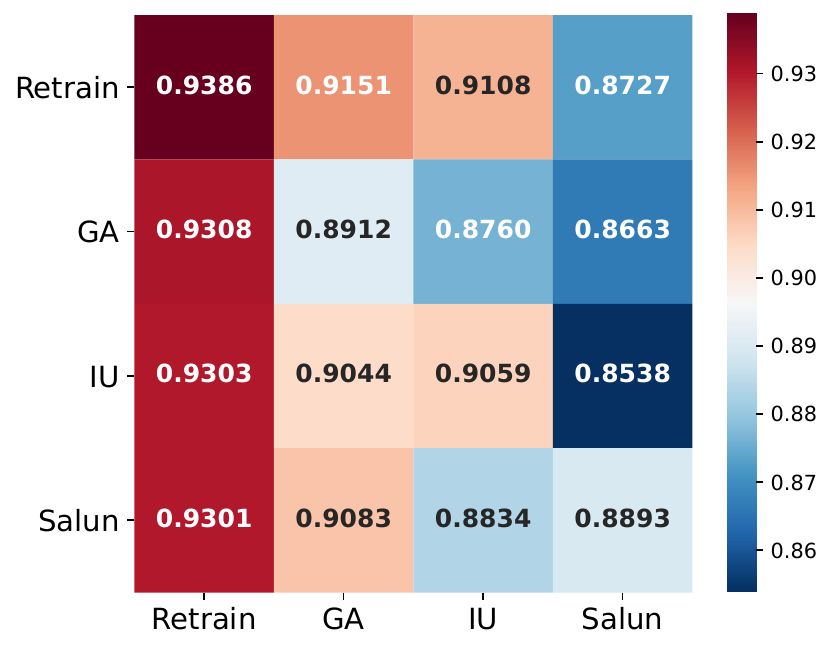}
        \caption{CIFAR10, ResNet18}
    \end{subfigure}%
    \hfill
    \begin{subfigure}[t]{0.23\linewidth}
        \centering
        \includegraphics[width=\linewidth]{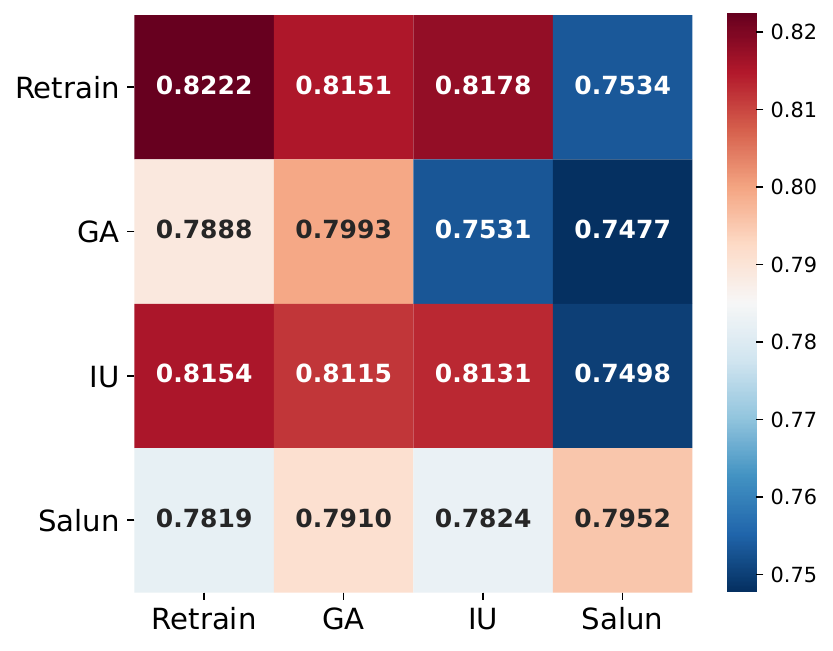}
        \caption{SVHN, ALLCNN}
    \end{subfigure}%
    \hfill
    \begin{subfigure}[t]{0.23\linewidth}
        \centering
        \includegraphics[width=\linewidth]{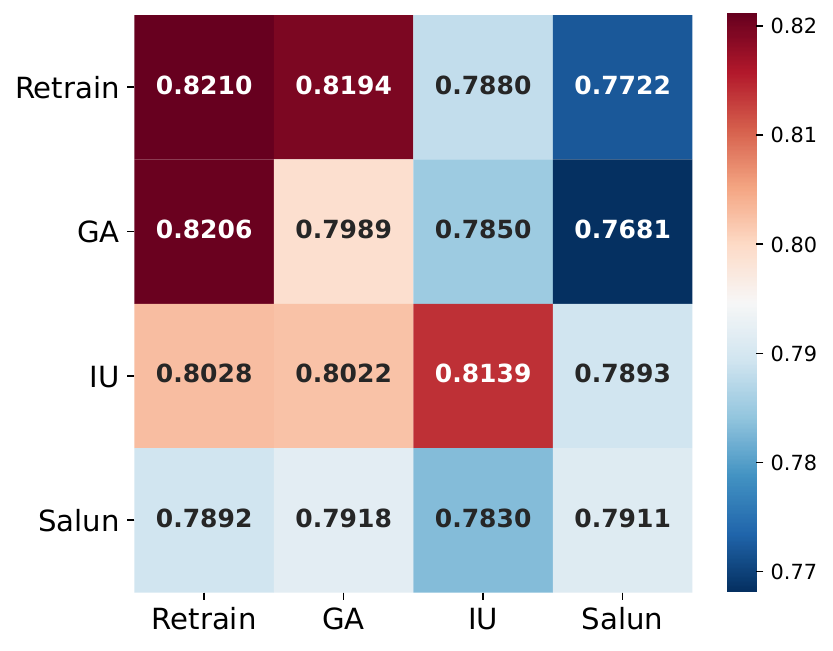}
        \caption{SVHN, ResNet18}
    \end{subfigure}

    \caption{Evaluation of the transferability of our method to different unlearning methods. The horizontal axis represents the unlearning settings of the shadow unlearned model, while the vertical axis represents the setting of the target model.}
    \label{fig:transferability_u}
\end{figure*}

\noindent \textbf{Transferability of Unlearning Methods.}  
We assess the transferability of our method across different unlearning methods. Figure~\ref{fig:transferability_u} shows that when both the shadow and target models use retraining, attack performance is optimal, as the shadow model closely simulates the target model. 
Our method remains effective even when the unlearning methods differ between the shadow and target models. This is because unlearning can create a notable distinction between the member and non-member UCD distributions. 
This indicates that even without knowing the unlearning algorithm of the target model, the attacker can still effectively infer the membership status of non-forgotten data.

\begin{figure}[t!]
    \centering
    \includegraphics[width=5cm]{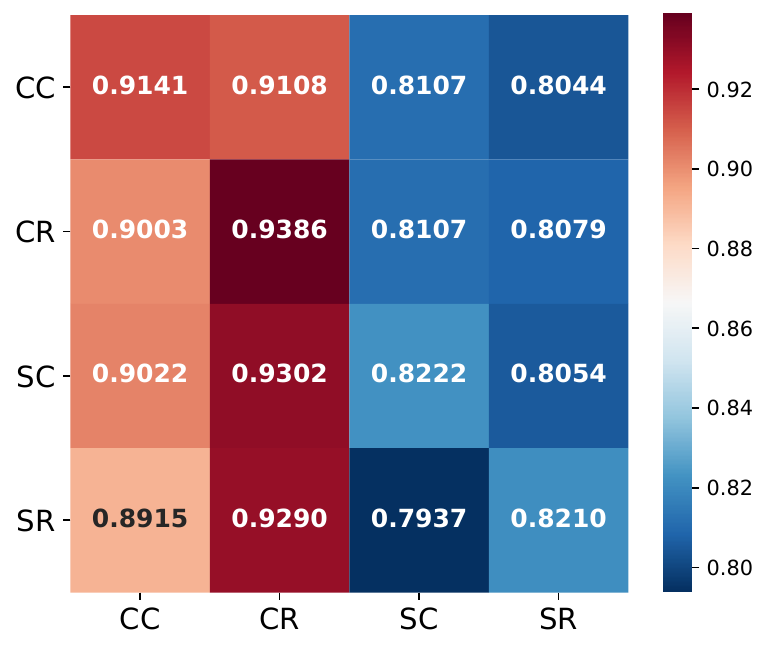}
\caption{ Analysis of transferability under different shadow model settings across datasets and architectures.}
    \label{fig:trans_data}
\end{figure}

\noindent Transferability of datasets and models.
Fig.~\ref{fig:trans_data} illustrates the strong cross-model and cross-dataset generalization ability of our method, which can be attributed to the significant distributional divergence between member and non-member samples.

\textbf{More Defenses.} We evaluate the performance of our method under early stopping defense, and the results are shown in Table~\ref{tab:es}. We chose to stop the model training at 10 and 50 epochs. It can be observed that early stopping struggles to strike a good balance between privacy protection and model performance.
\begin{table}[t]
  \centering

   \scalebox{0.8}{
    \begin{tabular}{lrrrr}
    \toprule
          & \multicolumn{2}{c}{10} & \multicolumn{2}{c}{50} \\
     \cmidrule(r){2-3}\cmidrule(r){4-5}
          & \multicolumn{1}{l}{MA} & \multicolumn{1}{l}{MIA} & \multicolumn{1}{l}{MA} & \multicolumn{1}{l}{MIA} \\
 \cmidrule(r){1-3}\cmidrule(r){4-5}
    CC    & 63.5200  & 0.8120  & 77.5600  & 0.8927  \\
    CR    & 62.3500  & 0.8175  & 67.3400  & 0.9166  \\
    SC    & 87.7000  & 0.6272  & 90.3400  & 0.7405  \\
    SR    & 85.4000  & 0.6766  & 89.3190  & 0.7447  \\
    \bottomrule
    \end{tabular}%
    }
      \caption{The performance of our method under early stopping, where MA (\%) and MIA represent the model accuracy and attack accuracy respectively.}
  \label{tab:es}%
\end{table}%

\begin{figure}[t!]
    \centering
    \begin{subfigure}[t]{0.48\linewidth}
        \centering
        \includegraphics[width=\linewidth]{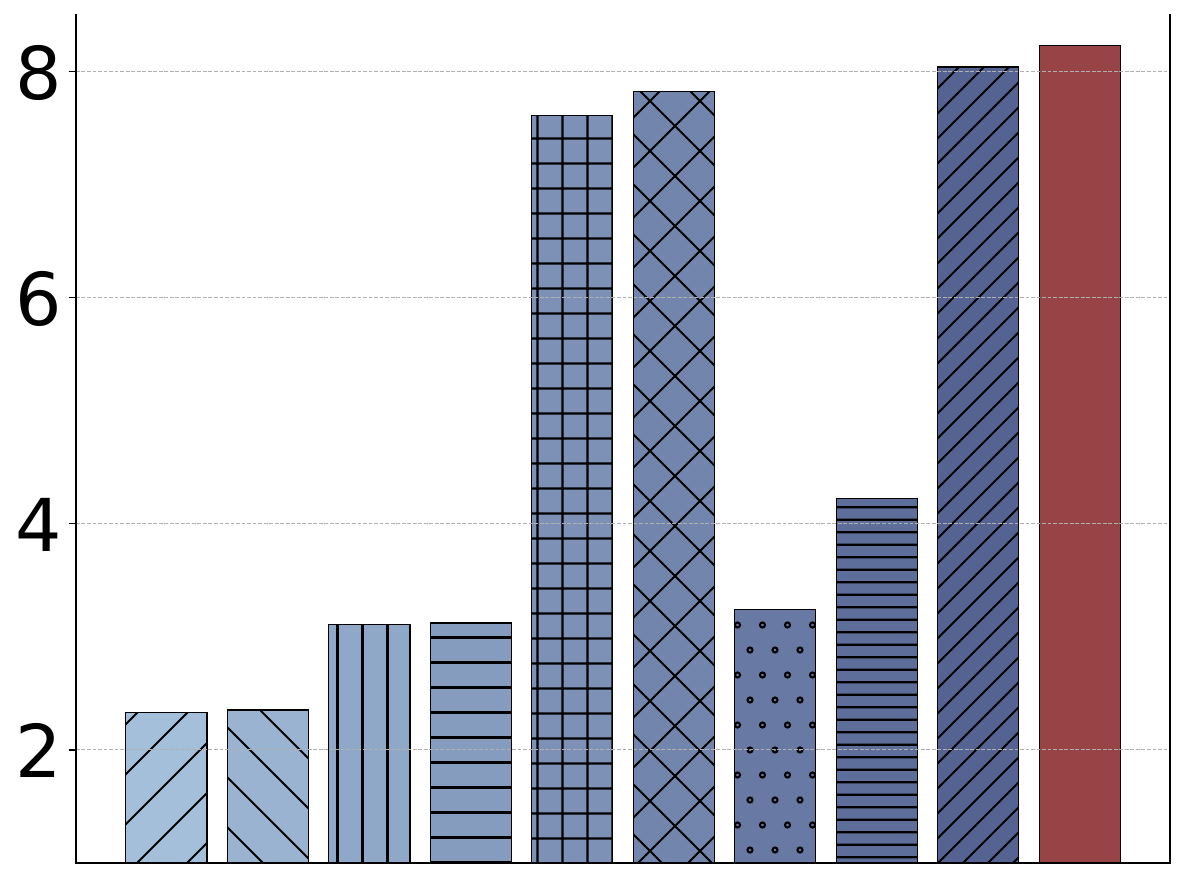}
        \caption{CIFAR10, ALLCNN}
    \end{subfigure}%
    \hfill
    \begin{subfigure}[t]{0.48\linewidth}
        \centering
        \includegraphics[width=\linewidth]{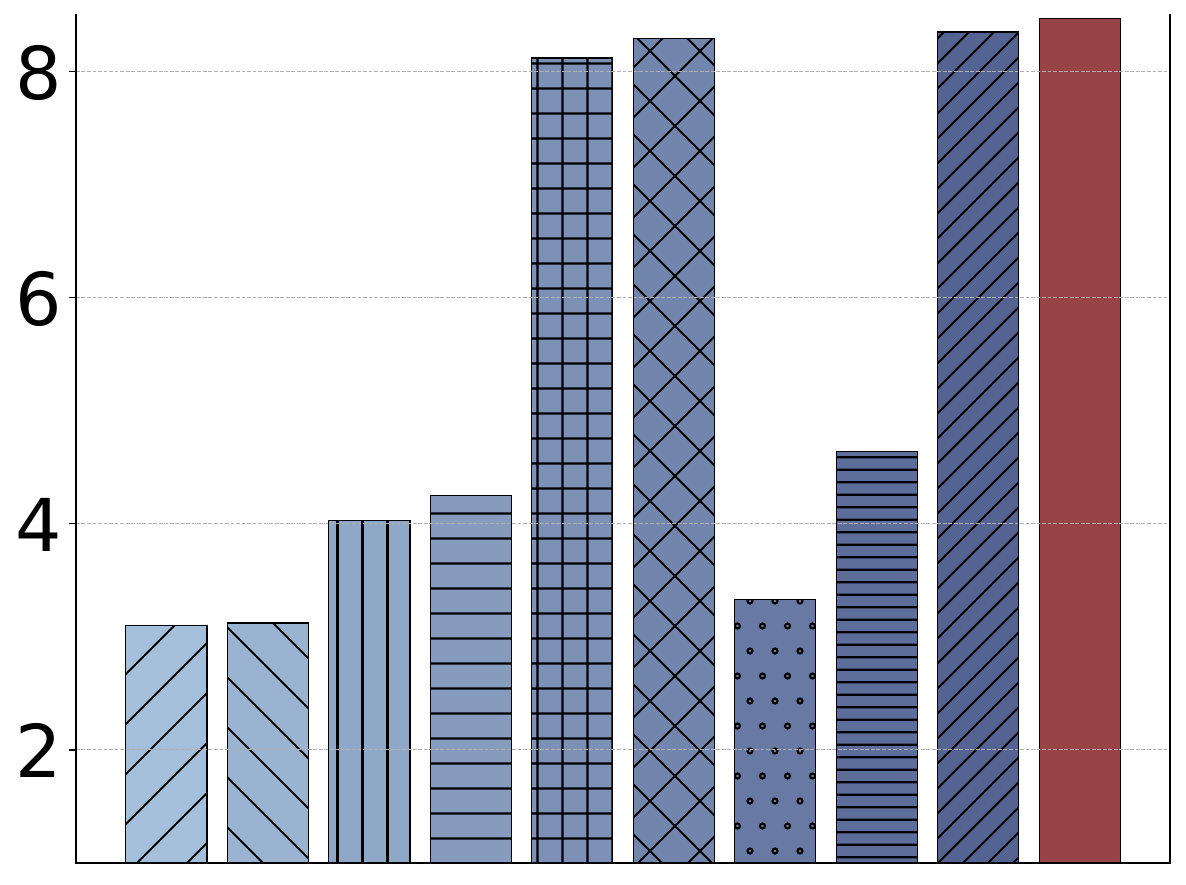}
        \caption{CIFAR10, ResNet18}
    \end{subfigure}
    
    \vspace{0.5em}  
    
    \begin{subfigure}[t]{0.48\linewidth}
        \centering
        \includegraphics[width=\linewidth]{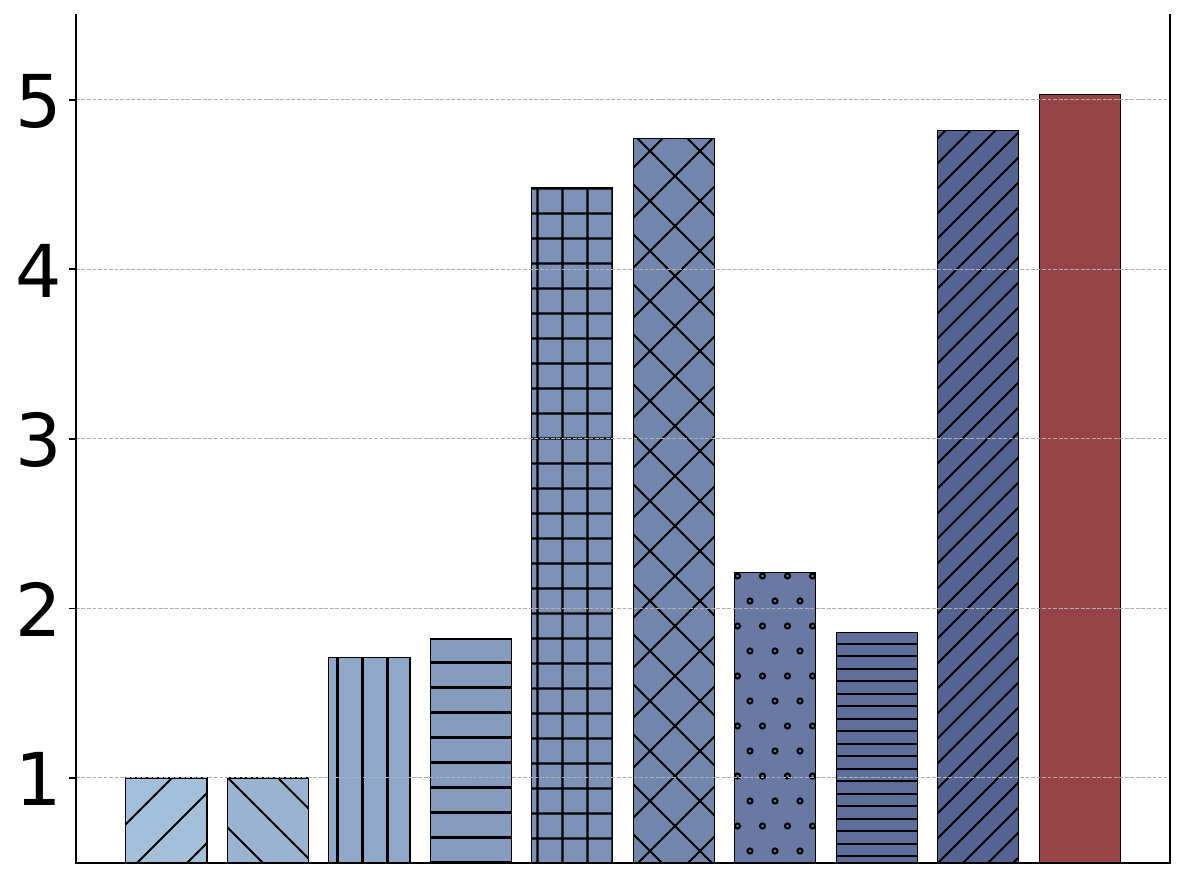}
        \caption{SVHN, ALLCNN}
    \end{subfigure}%
    \hfill
    \begin{subfigure}[t]{0.48\linewidth}
        \centering
        \includegraphics[width=\linewidth]{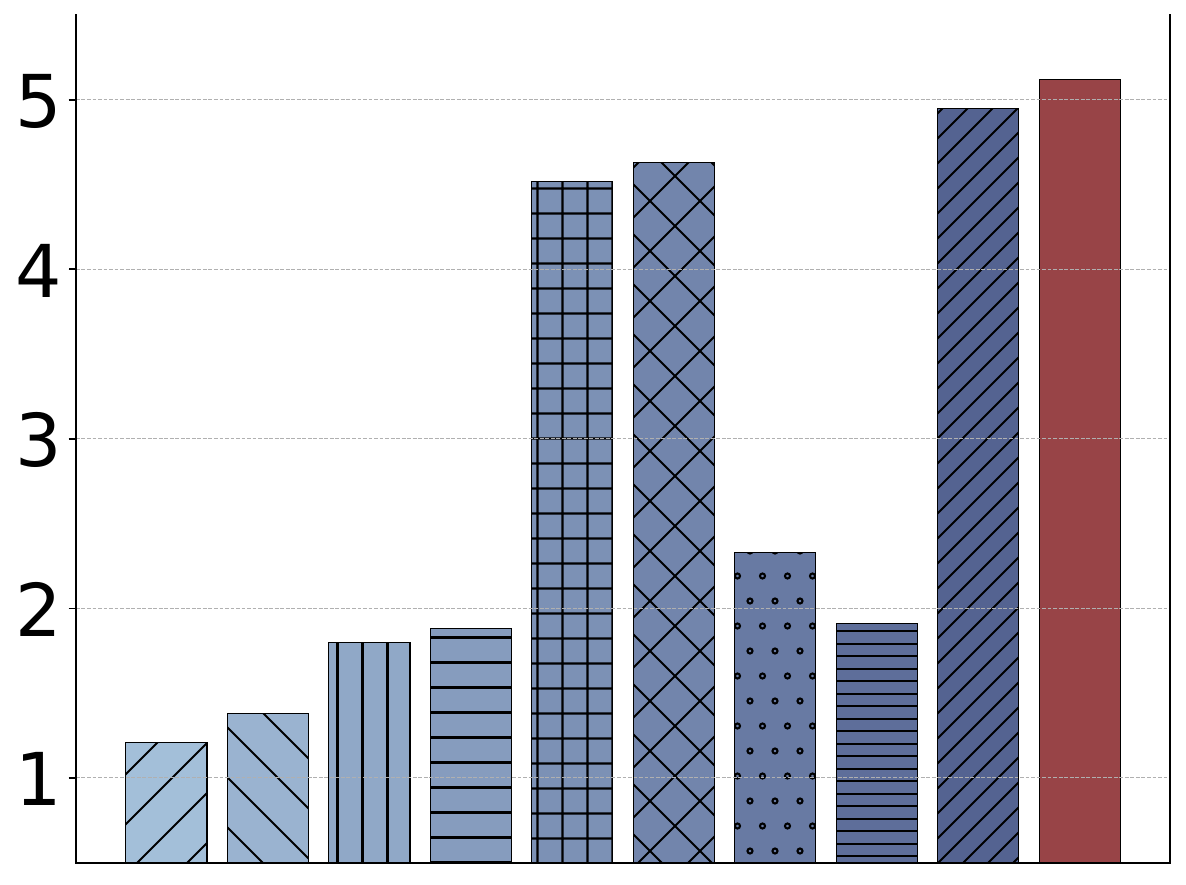}
        \caption{SVHN, ResNet18}
    \end{subfigure}
    \begin{subfigure}[t]{0.82\linewidth}
        \centering
        \includegraphics[width=\linewidth]{fig/method_legend.pdf}
        \caption{Legend illustration for different methods}
    \end{subfigure}
     \caption{Evaluation with TPR@0.1\%FPR.}
    \label{fig:tpr_comp}
\end{figure}

\textbf{Evaluation with TPR at low FPR.} As shown in Figure~\ref{fig:tpr_comp}, our method consistently achieves the best performance under this metric.
\subsection{Future Work}
Existing unlearning techniques seek to uphold users’ right to be forgotten by eliminating the influence of specific data from a trained model, thereby reducing the impact of {poisoned data}~\cite{zhou2025darkhash, badhash, wan2025mars, zhang2024detector, zhang2025test, wang2024trojanrobot, wang2024unlearnable, li2025detecting, wang2024eclipse, yu2025spa} on model behavior. However, the unlearning process can itself reveal behavioral differences between the model before and after unlearning, introducing new privacy risks. In this work, we are the first to identify such vulnerabilities in the retained data and provide an initial analysis of the associated risks. Our method relies on access to the model’s output probability distribution. In future work, we plan to explore a label‑only access setting~\cite{label_only,u-label}, where only predicted class labels are available, and incorporate {adversarial example techniques}~\cite{zhou2024securely, advclip, zhou2023downstream, zhou2025numbod, zhou2024darksam, zhou2025sam2, wang2025advedm, wang2025breaking, song2025segment, li2024transferable, song2025seg} to enable a more comprehensive and fine‑grained privacy evaluation.

\bibliography{aaai2026}
\end{document}